\documentclass[a4paper,fleqn]{cas-dc}

\usepackage[authoryear]{natbib}
\usepackage{algorithm}
\usepackage{algpseudocode}
\usepackage{graphicx}
\usepackage{siunitx}
\usepackage{multirow}
\usepackage{float}
\usepackage{tabularx}
\usepackage{array}      % 必须加载
\usepackage{booktabs}
\usepackage{stfloats}
\usepackage{placeins}
\usepackage{siunitx}
\def\tsc#1{\csdef{#1}{\textsc{\lowercase{#1}}\xspace}}
\tsc{WGM}
\tsc{QE}
\newcommand{\R}{\mathbb{R}}
\newcommand{\G}{\mathcal{G}}
\newcommand{\V}{\mathcal{V}}
\newcommand{\E}{\mathcal{E}}
\newcommand{\M}{\mathcal{M}}

\newcommand{\Lc}{\mathcal{L}}
\newcommand{\Ac}{\mathbf{A}}
\newcommand{\X}{\mathbf{X}}
\newcommand{\Hh}{\mathbf{H}}

\newcommand{\eps}{\varepsilon}

\begin{document}
\let\WriteBookmarks\relax
\def\floatpagepagefraction{1}
\def\textpagefraction{.001}

% Short title
\shorttitle{}    

% Short author
\shortauthors{}  

% Main title of the paper
\title [mode = title]{ArborNet: Volume- and skeleton-aware completion of tree quantitative structure models from incomplete point clouds using anchor-conditioned graph learning}  

% Title footnote mark
% eg: \tnotemark[1]
%\tnotemark[1] 

% Title footnote 1.
% eg: \tnotetext[1]{Title footnote text}
%\tnotetext[1]{This research was supported by the General Research Fund (Project 15215023) by the Research Grants Council of the Hong Kong Special Administrative Region.} 
\nonumnote{This research was supported by the General Research Fund (Project 15215023) by the Research Grants Council of the Hong Kong Special Administrative Region.} 

% First author
%
% Options: Use if required
% eg: \author[1,3]{Author Name}[type=editor,
%       style=chinese,
%       auid=000,
%       bioid=1,
%       prefix=Sir,
%       orcid=0000-0000-0000-0000,
%       facebook=<facebook id>,
%       twitter=<twitter id>,
%       linkedin=<linkedin id>,
%       gplus=<gplus id>]

\author[1]{Ge Jiang}[orcid=0000-0001-9192-5444]

% Footnote of the first author
%\fnmark[1]

% Email id of the first author
\ead{20034646r@connect.polyu.hk}

% URL of the first author
%\ead[url]{}

% Credit authorship
% eg: \credit{Conceptualization of this study, Methodology, Software}
%\credit{}

\author[1, 2, 3, 4]{Tiangang Yin}[type=editor, orcid=0000-0002-2149-6004]

% Corresponding author indication
\cormark[1]
\ead{tiangang.yin.85@gmail.com}

\author[5, 6]{Wei Yao}[orcid=0000-0001-7704-0615]
\cormark[1]
\ead{wyao@iue.ac.cn}

\author[7]{Wai Yeung Yan}[orcid=0000-0001-7813-1666]
\ead{waiyeung.yan@torontomu.ca}

\author[1]{Qi Chen}[orcid=0000-0002-5975-9092]
\ead{qi-chee.chen@connect.polyu.hk}

\author[1]{Jinhong Wang}[orcid=0000-0001-6314-0906]
\ead{jinhong0229.wang@connect.polyu.hk}

\author[1, 2, 3, 4]{Qihao Weng}[orcid=0000-0002-2498-0934]
\ead{qihao.weng@polyu.edu.hk}

% Address/affiliation
\affiliation[1]{organization={Department of Land Surveying and Geospatial Science, The Hong Kong Polytechnic University},
	city={Hung Hom},
	country={Hong Kong}
}

\affiliation[2]{organization={JC STEM Lab of Earth Observations, The Hong Kong Polytechnic University},
	city={Hung Hom},
	country={Hong Kong}
}

\affiliation[3]{organization={Research Centre for Artificial Intelligence in Geomatics, The Hong Kong Polytechnic University},
	city={Hung Hom},
	country={Hong Kong}
}

\affiliation[4]{organization={Research Institute for Land and Space, The Hong Kong Polytechnic University},
	city={Hung Hom},
	country={Hong Kong}
}

\affiliation[5]{organization={Spatial Intelligence and Urban Computing, Institute of Urban Environment, Chinese Academy of Sciences},
            city={Xiamen},
            country={China}}
            
\affiliation[6]{organization={State Key Lab for Ecological Security of Regions and Cities, Institute of Urban Environment, CAS},
%	addressline={}, 
	city={Xiamen},
	%          citysep={}, % Uncomment if no comma needed between city and postcode
%	postcode={}, 
%	state={},
	country={China}
	}
	
\affiliation[7]{organization={Department of Civil
		Engineering, Toronto Metropolitan University},
	%	addressline={}, 
	city={Toronto},
	%          citysep={}, % Uncomment if no comma needed between city and postcode
	%	postcode={}, 
		state={Ontario},
	country={Canada}
}

% Corresponding author text
\cortext[1]{Corresponding author}

% Footnote text
%\fntext[1]{}

% For a title note without a number/mark
%\nonumnote{}

% Here goes the abstract
\begin{abstract}
			Terrestrial laser scanning (TLS) provides detailed three-dimensional observations of individual trees and enables reconstruction of quantitative structure models (QSMs) for estimating woody volume and structural traits. However, single-station or sparsely sampled TLS acquisitions are frequently affected by occlusion, leading to incomplete branch observations and biased QSM-derived volume estimates. Existing point cloud completion methods can generate visually plausible shapes, but they do not explicitly preserve branch topology, cylinder radii, or woody volume, which are essential for biomass estimation and ecological applications. This paper presents a volume- and skeleton-aware QSM completion framework for recovering complete tree woody structures from incomplete TLS-derived QSMs. Rather than completing raw point clouds, the proposed method first represents each tree as a TreeQSM cylinder graph, where each skeleton node corresponds to a cylinder parameterized by its start position, axis, length, radius, branch level, parent--child topology, and branch identity. A multi-relation graph encoder propagates information along parent, child, and intra-branch edges. An anchor-conditioned decoder then predicts missing cylinders through two complementary generation modes: terminal branch continuation and new branch generation. To improve structural and biophysical consistency, the network is trained with geometry, existence, branch-start, branch-level, radius-prior, axis-smoothness, total-volume, and level-wise-volume losses. During inference, the framework accepts generated cylinders using predicted level-wise missing volume budgets. The framework is designed for both simulated TLS tree data generated by the DART radiative transfer model and real urban TLS trees collected in Hong Kong, where single-station QSMs are treated as partial observations and multi-station QSMs as complete references. We evaluated ArborNet on DART-simulated TLS data and urban TLS trees acquired in Hong Kong. Relative to the selected reference QSMs, the proposed method achieved a mean total-volume relative error of 0.081\% and a mean level-wise volume error of 0.676\% under predicted-budget-constrained inference. These values quantify agreement with the reference QSMs rather than accuracy relative to destructively measured woody volume. The proposed method also reduced skeleton geometry error compared with the evaluated point-cloud completion baselines. Ablation studies support the contribution of the evaluated loss terms to reference-volume agreement, with joint volume supervision being particularly important for volume control. This study highlights QSM-domain completion as an alternative to point-cloud completion for jointly modelling explicit woody structure and volume, extending tree completion research beyond geometric shape recovery.
\end{abstract}

% Use if graphical abstract is present
%\begin{graphicalabstract}
%\includegraphics{}
%\end{graphicalabstract}

% Research highlights
\begin{highlights}
\item A QSM-domain tree completion framework is proposed for incomplete TLS-derived woody structures.
\item Level-wise missing woody volume is explicitly predicted and used as a constrained decoding budget.
\item TreeQSM cylinders are converted into a skeleton-aware skeleton graph with geometric and structural attributes.
\item Missing cylinders are generated by anchor-conditioned continuation and branch-start decoders.
\item Completion predicts connected QSM cylinders with explicit geometry and woody volume.
\end{highlights}

%\nocite{*}

% Keywords
% Each keyword is seperated by \sep
\begin{keywords}
 			LiDAR point cloud \sep Quantitative structure models \sep Tree skeleton completion  \sep Graph neural network \sep Urban trees \sep DART
\end{keywords}

\maketitle
\section{Introduction}
% ============================================================
Woody structures of trees store a large fraction of terrestrial carbon, and their accurate quantification is essential for carbon accounting and climate change mitigation \citep{pan2011large, chen2015modeling, harris2021global}. Traditionally, tree volume and above-ground biomass have been estimated using allometric models that relate easily measurable field attributes, such as diameter at breast height (DBH) and tree height, to volume or biomass through power-law relationships \citep{segura2005allometric, bhebhe2025estimation, ali2026multisource}. These models are widely applied in forest inventories as they are simple, inexpensive, and generally sufficient for large-scale carbon accounting when detailed tree-level volume measurements are unavailable \citep{chave2005tree}. However, allometric models provide only aggregate statistical estimates and do not resolve branch-level architecture, topology, or the spatial distribution of woody mass \citep{dong2025remote, honkanen2026transferring}. As a result, they are not intended for detailed 3D structural analysis, visualization, or applications that require explicit tree geometry. Moreover, their accuracy is species-specific and site-dependent, often introducing substantial uncertainty when applied beyond their calibration range \citep{lockwood2021assessing, ledo2016re}.

Terrestrial laser scanning (TLS) has emerged as a major technique that overcomes these limitations by directly characterizing tree architecture at the individual-tree scale \citep{li2025terrestrial, bornand2023individual, polewski2017voting}. Dense TLS point clouds allow detailed reconstruction of stems and branches, from which quantitative structure models (QSMs) can be derived to estimate woody volume, branch size distribution, tree architecture, and above-ground biomass proxies \citep{lau2018quantifying, gonzalez2018estimation, morales2025reducing}. Beyond visual reconstruction, explicit 3D tree models are increasingly required as geometric inputs for radiative transfer and canopy reflectance modelling \citep{widlowski2014abstract}, photosynthesis and 3D light-interception simulations \citep{wu2022canopy}, and computational studies of canopy airflow and turbulence \citep{ahmed2026spatial}. Such applications depend not only on the total woody volume but also on branch topology, branch orientation, and the spatial arrangement of woody elements, which cannot be provided by allometric estimates. QSM-based tree modeling is particularly valuable as it converts unstructured point clouds into an explicit collection of connected geometric primitives, usually cylinders, that represent the woody skeleton of a tree \citep{ali2025hybrid, ye2019analysing}.

Despite these advantages, TLS observations are often incomplete \citep{morhart2024limitations}. Occlusion caused by stems, branches, leaves, neighbouring trees, urban objects, and scanner-view geometry can result in missing branches or truncated branch tips \citep{boucher2021sampling, wan2019quantification}. Multi-station TLS can alleviate this problem, but it requires longer field acquisition time, more complicated registration, and higher processing cost. In many practical forest, urban, and ecological surveys, only one or a few scan positions are available. As a result, QSMs reconstructed from partial TLS point clouds frequently underestimate branch number, branch length, and woody volume \citep{martin2021evaluation, dai2022comparison, liu2018estimating}. Such incompleteness directly limits the reliability of QSM-derived structural and biomass estimates.

In light of these limitations, 3D completion offers a promising direction. Recent deep learning methods for point cloud completion, including PoinTr \citep{yu2021pointr}, AdaPoinTr \citep{10232862}, and SnowflakeNet \citep{xiang2021snowflakenet}, have achieved impressive performance for general object shape completion. These methods typically learn to generate dense point sets from incomplete point clouds using Chamfer distance, Earth Mover's Distance, or related geometric losses \citep{zhang2024deep, fei2022comprehensive}. Several recent studies have extended these techniques to tree point clouds. \citet{bornand2024completing} fine-tuned PoinTr on synthetic and real TLS data to fill gaps in 1 m\(^3\) tree point cloud segments. \citet{bornand2026reconstructing} further adapted AdaPoinTr to reconstruct coniferous crown alpha-shapes for height estimation. \citet{xindi2025advanced} enhanced SnowflakeNet with a global context module and channel attention for urban tree completion from ALS and MLS data. SK-TreePCN \citep{xu2025sk} introduced skeleton-embedded point cloud completion, using tree skeleton topology as a structural prior to guide the completion process.

Despite these advances, tree woody structure completion differs from generic shape completion in several important aspects \citep{livny2010automatic, aakerblom2015analysis}. First, the target is not merely a visually plausible surface or point distribution, but a biologically meaningful branching structure \citep{born2016branched}. Second, the recovered structure must preserve parent--child topology, branch order, tapering, and geometric continuity \citep{raumonen2013fast, smith2014tree}. Third, for ecological applications, the most important quantity is often woody volume rather than visual shape \citep{macfarlane2026deconstructing, dassot2012terrestrial}. A point cloud completion result that appears visually complete may still yield inaccurate QSM reconstruction and biased volume estimates \citep{demol2022volumetric, morhart2024limitations, abegg2023tree}. Moreover, all of the above methods operate at the point-cloud level: they complete raw points rather than structural models. Their outputs must be fed into a separate QSM reconstruction pipeline, e.g., TreeQSM \citep{raumonen2013fast, calders2018realistic, brede2019non, calders2022laser}, to obtain cylinder-based tree models, which introduces additional uncertainty and may lose topological information during the conversion process.

To address these limitations, we propose a QSM-domain completion framework that directly operates on cylinder skeletons derived from TreeQSM \citep{calders2018realistic, calders2022laser} rather than raw point clouds. Each QSM cylinder is treated as a skeleton node containing geometric and topological attributes, including start point, end point, centre, axis, length, radius, branch level, branch-start indicator, topological depth, branch-relative position, parent radius ratio, and turning angle. The partial tree is represented as a multi-relation graph with parent, child, and intra-branch adjacency matrices. A graph neural encoder extracts skeleton-aware node embeddings, and an anchor-conditioned decoder predicts missing cylinders associated with observed cylinders.

The central idea is that missing tree structures can be decomposed into two dominant modes, namely continuation of observed terminal branches and emergence of missing side branches from observed internal branches. Accordingly, the proposed decoder contains a continuation head and a branch-start head. The continuation head predicts a short chain of missing cylinders from each observed tip, while the branch-start head predicts longer branch chains from eligible non-terminal anchors. In addition, the network predicts the amount of missing woody volume at each branch level. This level-wise volume prediction is used during inference as a generation budget, ensuring that accepted cylinders are not only locally plausible but also globally consistent with expected missing volume.

The main contributions of this work are as follows:
\begin{enumerate}
	\item We formulate incomplete tree reconstruction as a QSM skeleton completion problem rather than a raw point cloud completion problem, enabling explicit prediction of branch topology, radius, length, and woody volume.
	\item We develop a skeleton-aware multi-relation graph encoder and an anchor-conditioned dual-mode decoder using explicit parent–child relationships, and combine volume supervision with predicted-budget-constrained decoding.
	\item We establish a reference-aligned QSM completion evaluation using simulated Discrete Anisotropic Radiative Transfer (DART) \citep{gastellu2016simulation} TLS data and real Hong Kong urban TLS trees, with observed regions identified from partial scans and complete-QSM references obtained from complete simulated or registered multi-station point clouds.
\end{enumerate}

The remainder of this paper is organized as follows. Section \ref{sec:2} reviews related work. Section \ref{sec:3} describes the proposed methodology. Section \ref{sec:4} details the experimental setup. Section \ref{sec:5} presents results. Section \ref{sec:6} discusses implications and limitations. Section \ref{sec:7} concludes the paper.

% ============================================================
\section{Related work} \label{sec:2}
% ============================================================

\subsection{TLS-based tree structure modelling}

TLS has been widely used to capture three-dimensional tree structure. Compared with airborne laser scanning (ALS), TLS provides substantially higher point density and can capture detailed stem and branch geometry \citep{opitz2013overview, crespo2020comparative, fekry2022ground, yao2012tree}. From TLS point clouds, QSM methods reconstruct trees as collections of connected cylinders or tapered cylinders. Such models have been used to estimate woody volume, biomass, branch architecture, canopy structure, and ecological traits. TreeQSM \citep{raumonen2013fast, calders2018realistic, calders2022laser} is among the widely used approaches for reconstructing cylinder-based tree models from woody point clouds. Other notable QSM methods include SimpleTree \citep{hackenberg2015simpletree}, AdQSM \citep{fan2020adqsm}, AdTree \citep{du2019adtree}, Treegraph \citep{yang2024treegraph}, etc. More recently, SmartQSM \citep{yang2026smartqsm} introduced a deep learning-based approach that uses sparse-convolution-based point cloud contraction to directly reconstruct QSMs from raw TLS point clouds, achieving state-of-the-art accuracy in trunk volume and branch parameter extraction.

Nevertheless, the quality of QSMs strongly depends on point cloud completeness \citep{pyorala2018quantitative, chen2025advancing, ye2019analysing}. Occluded branches, low point density, and imperfect wood--leaf separation can lead to missing or inaccurate reconstruction. Multi-scan acquisition can improve coverage, but single-scan or sparse-scan TLS remains common in operational scenarios \citep{kelbe2015single, wallace2022terrestrial}. SmartQSM \citep{yang2026smartqsm}, similar to other QSM reconstruction methods, assumes that the input point cloud is sufficiently complete; it does not address the problem of completely missing branches that are not captured in the original TLS acquisition. These gaps motivate the development of completion methods specifically designed for tree structures, which we review in the following subsections.

\subsection{Tree point cloud completion}

Deep learning-based point cloud completion has developed rapidly \citep{zhang2024deep, tesema2023point, fei2022comprehensive}. Encoder--decoder networks \citep{yuan2018pcn, yu2021pointr, chen2024learning}, transformer-based methods \citep{yu2021pointr, 10232862}, folding-based decoders \citep{yang2018foldingnet, li2024carvingnet}, and coarse-to-fine refinement networks \citep{liu2020morphing, chang2021finerpcn} have been proposed to infer complete point clouds from partial observations. PoinTr \citep{yu2021pointr} formulates point cloud completion as a set-to-set translation problem using transformer architectures. AdaPoinTr \citep{10232862} further improves adaptability and representation learning. SnowflakeNet \citep{xiang2021snowflakenet} progressively generates point clouds using snowflake point deconvolution.

Several studies have extended these methods to tree point clouds. \citet{bornand2024completing}  conducted a comprehensive analysis of training data simulation strategies for tree point cloud completion, fine-tuning the transformer-based PoinTr model on both synthetic and real TLS data to fill gaps in 1 m\(^3\) segments of tree point clouds. \citet{bornand2026reconstructing} further adapted AdaPoinTr to reconstruct the outer crown shape (alpha-shape hull) of coniferous trees from incomplete point clouds, demonstrating improved height estimation accuracy on three independent datasets. \citet{xindi2025advanced} enhanced SnowflakeNet with a global context module and channel attention mechanisms for urban tree completion from ALS and MLS data, achieving substantial reductions in completion errors.

Although these methods can produce visually plausible complete point clouds, they share a common limitation in that they operate at the point-cloud level and are not specifically designed for QSM-derived structural traits. In tree applications, a visually plausible point distribution may still lack correct branch connectivity, radius tapering, and volume. Moreover, downstream QSM reconstruction from completed point clouds introduces another source of uncertainty, as the conversion from points to cylinders may further distort topological relationships \citep{disney2018weighing, calders2022laser, demol2022volumetric}. This motivates completion directly in the QSM representation space.

\subsection{Graph learning for tree and skeleton structures}

Trees are naturally represented as graphs, with nodes corresponding to branch segments or skeleton points and edges corresponding to topological connectivity \citep{jiang2021skeleton, gao2025extraction, hu2017efficient}. Graph neural networks provide a suitable framework for learning from such structured representations \citep{li2025graph, fischer2021stickypillars, fekry2023individual}. Unlike point-based networks that treat samples as unordered sets, graph networks can explicitly exploit parent--child and branch-continuity relations \citep{yan2025connectivity}. For QSM completion, graph learning is particularly appropriate because each cylinder already contains geometric, topological, and branch-level information \citep{kulicki2024artificial, xu2025sk, zhou2025treestructor}. 

SK-TreePCN \citep{xu2025sk} is a representative example that leverages skeleton graphs to guide point cloud completion, where the tree skeleton topology is extracted as a structural prior to guide the completion process through a skeleton proxy point sampler and a skeleton topology adjuster, effectively reducing point cloud dispersion and discontinuity issues. However, the method still operates at the point-cloud level, requiring a separate QSM reconstruction step after completion. This suggests that integrating topology as an inherent part of the representation, rather than as an external constraint, may be a promising direction for tree completion.

The task is also related to conditional graph generation and neural generation of hierarchical structures. General graph generators such as GraphRNN \citep{you2018graphrnn} and GRANs \citep{liao2019efficient} generate graph nodes and edges sequentially or blockwise, whereas relational graph neural networks such as R-GCN \citep{schlichtkrull2018modeling} distinguish multiple edge types during message passing. Recursive \citep{guo2019dynamic} and tree-structured decoders \citep{alvarez2017tree} further model hierarchical dependencies. However, these unconstrained graph generation is limited to graph structure without considering jointly predicting continuous cylinder geometry, radius, and volume of tree QSMs.

\subsection{Volume-aware structural completion}

Most generic point cloud completion methods optimize geometric distances such as Chamfer distance or Earth Mover's Distance \citep{wu2021density, wu2021balanced, li2026flexible}. However, for forest and urban tree applications, woody volume is often a primary target variable. A small geometric error in large-radius branches may cause substantial volume error, whereas visually noticeable errors in small twigs may have limited volume impact \citep{franke2023vet, morales2025rtwig}. These considerations motivate losses that constrain both total woody volume and its distribution across branch levels.

\subsection{Procedural and biologically informed tree generation}
Procedural and biologically informed tree modeling, including L-systems \citep{allen2005using}, space-colonization \citep{runions2007modeling}, and inverse procedural modeling \citep{stava2014inverse} can infer missing woody structures and regularize branch dimensions using biological priors like the pipe-model theory and allometry; however, these methods face limitations when completing partial LiDAR observations. Their performance heavily relies on manually specified or species-specific parameters that generalize poorly across diverse growth environments and age classes. Furthermore, inverse parameter estimation remains highly underdetermined when major branches are occluded, and idealized growth rules struggle to capture capture irregular architectures caused by pruning, damage, or urban conditions. Ultimately, because biological constraints primarily regulate aggregate dimensions rather than uniquely determining the spatial position and connectivity of unobserved branches, these methods often generate biologically plausible trees that fail to accurately reconstruct the specific structure of the reference observation.

% ============================================================
\section{Methodology} \label{sec:3}
% ============================================================
For reference, Table~\ref{tab:symbols} summarizes the key symbols used throughout the methodology section. Symbols are grouped by their primary context of use.

\begin{table*}[htbp]
	\centering
	\caption{Summary of key symbols used in the methodology.}
	\label{tab:symbols}
	\begin{tabular}{p{4cm} p{1.5cm} p{9cm}}
		\toprule
		\textbf{Category} & \textbf{Symbol} & \textbf{Description} \\
		\midrule
		\multirow{4}{*}{QSM cylinder graph} 
		& $\G^\mathrm p$ & Partial QSM cylinder graph \\
		& $\G^\mathrm r$ & Reference QSM cylinder graph \\
		& $\hat{\G}^\mathrm c$ & Completed QSM cylinder graph (predicted) \\
		& $\M$ & Set of missing cylinders to be attached to $\G^\mathrm p$ \\
%		& $V(\G)$ & Total volume of all cylinders in graph $\G$ \\
%		& $V_\ell(\G)$ & Volume contributed by cylinders at level $\ell$ in graph $\G$ \\
		\midrule
		\multirow{11}{*}{Cylinder attributes} 
		& $\mathbf{q}_i$ & Raw attribute vector of cylinder $i$ (13 columns) \\
		& $\mathbf{c}_i$ & 8-dimensional geometry vector of cylinder $i$ \\
		& $\mathbf{x}_i$ & 20-dimensional input feature vector of cylinder $i$ \\
		& $\mathbf{s}_i$ & Start point of cylinder $i$ ($\in \R^3$) \\
		& $\mathbf{e}_i$ & End point of cylinder $i$ ($\in \R^3$) \\
		& $\mathbf{o}_i$ & Centre point of cylinder $i$ ($\in \R^3$) \\
		& $\mathbf{a}_i$ & Axis direction of cylinder $i$ ($\in \R^3$, unit vector) \\
		& $r_i$ & Radius of cylinder $i$ ($\in \R^+$) \\
		& $l_i$ & Length of cylinder $i$ ($\in \R^+$) \\
		& $\tilde{\ell}_i$ & Normalized branch level of cylinder $i$ ($\in \{0,1,2,3\}$) \\
		& $b_i$ & Branch identifier of cylinder $i$ ($\in \mathbb{N}$) \\
		%		& $p_i$ & Parent identifier of cylinder $i$ \\
		%		& $\gamma_i^{TLS}$ & Normalized TLS scan count ratio ($c_{\text{TLS}}/3$) \\
		\midrule
		\multirow{6}{*}{QSM graph construction} 
		& $\V^p$ & Set of nodes in the partial graph (each node is a cylinder) \\
		& $\X$ & Node feature matrix ($N \times 20$) \\
		& $\Ac_\mathrm p$ & Parent adjacency matrix \\
		& $\Ac_\mathrm c$ & Child adjacency matrix \\
		& $\Ac_\mathrm b$ & Intra-branch adjacency matrix \\
		& $\tilde{\Ac}_{ij}$ & Row-normalized adjacency entry \\
		\midrule
		\multirow{13}{*}{Encoder and decoder} 
		& $\Hh^{(t)}$ & Node embedding matrix at layer $t$ ($N \times 128$) \\
		& $\mathbf{Z}$ & Final encoder output (node memory embeddings, $N \times 128$) \\
		& $\mathbf{z}_i$ & Encoder output for node $i$ ($\in \R^{128}$) \\
		& $\mathbf{M}^{(t)}$ & Aggregated message matrix at layer $t$ ($N \times 128$) \\
		& $\hat{\mathbf{s}}_{i,k}$ & Predicted start point of cylinder $k$ at anchor $i$ \\
		& $\hat{\mathbf{a}}_{i,k}$ & Predicted axis direction of cylinder $k$ at anchor $i$ \\
		& $\hat{l}_{i,k}$ & Predicted length of cylinder $k$ at anchor $i$ \\
		& $\hat{r}_{i,k}$ & Predicted radius of cylinder $k$ at anchor $i$ \\
		& $\hat{z}_{i,k}$ & Existence logit of cylinder $k$ at anchor $i$ \\
		& $\hat{z}^\mathrm{start}_{i}$ & Branch-start logit at anchor $i$ \\
		& $\eta_i$ & Target missing-volume budget at anchor $i$ ($\in \mathbb{R}$) \\
		& $\hat{\eta}_i$ & Predicted missing-volume budget at anchor $i$ \\
		& $\hat{\mathbf{V}}^\mathrm{global}$ & Predicted global level-wise volume budget ($\in \R^4$) \\
		\midrule
		\multirow{5}{*}{Training} 
		& $\Omega$ & Set of all anchor-slot pairs $(i,k)$ \\
		& $\mathcal{P}$ & Set of positive anchor-slot pairs (cylinders exist in reference QSM) \\
		& $\Omega_\mathrm b$ & Set of anchors eligible for branch-start generation \\
		& $\Lc$ & Total multi-task loss \\
		& $\lambda_{(\cdot)}$ & Loss weights for individual loss components \\
		\bottomrule
	\end{tabular}
\end{table*}

\subsection{Overview}

The proposed framework completes partial tree structures in the QSM domain. Given a partial QSM graph \(\G^p\), the objective is to predict a set of missing cylinders \(\hat{\M}\) such that the completed QSM
\begin{equation}
	\hat{\G}^c = \G^p \cup \hat{\M}
\end{equation}
approximates the reference QSM \(\G^r\) in terms of geometry, topology, branch-level structure, and woody volume.

\begin{figure*}[!ht]
	\centering
	\includegraphics[width=1\linewidth]{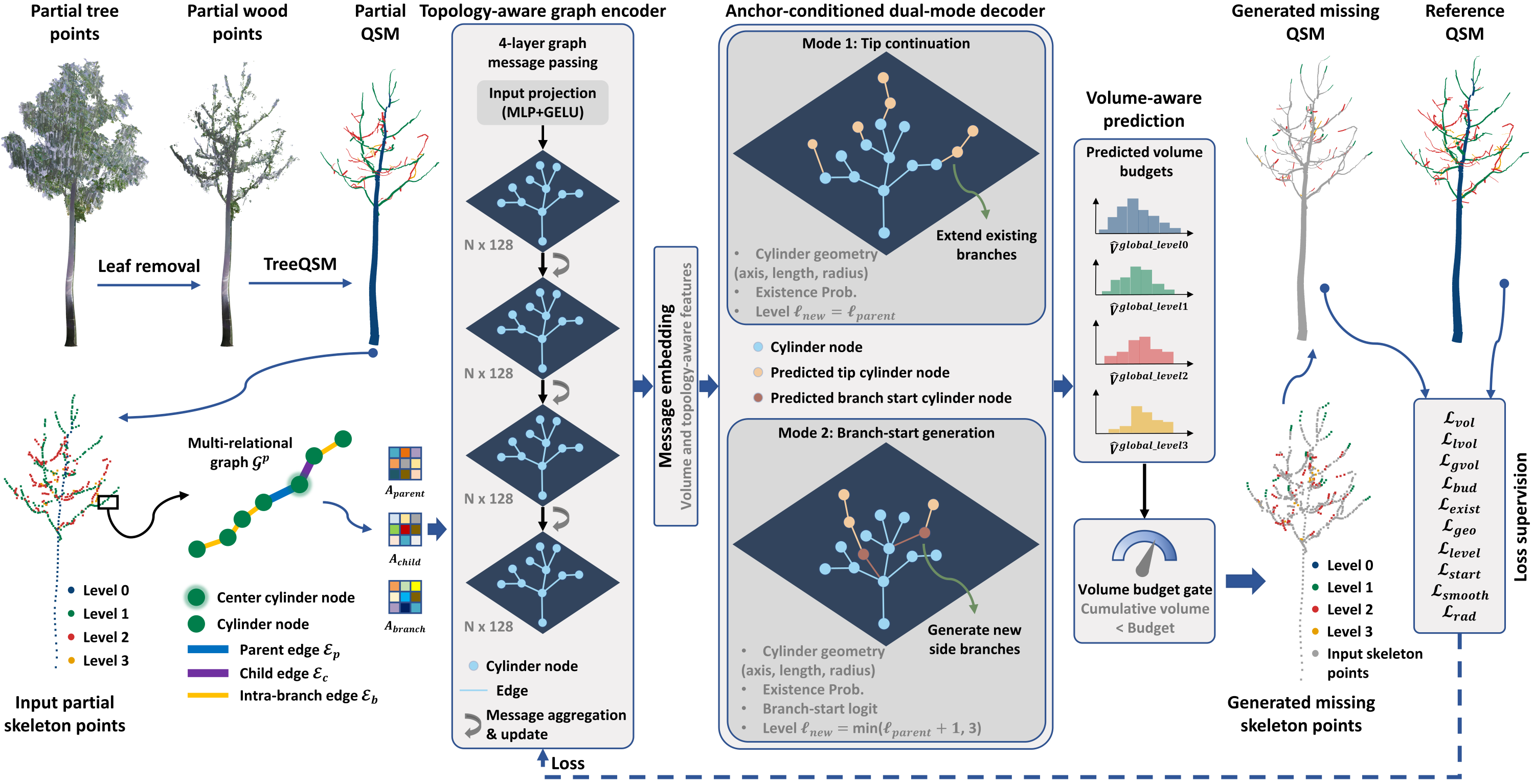}
	\caption{Overview of the proposed QSM-domain tree completion framework. The pipeline consists of four stages: (i) partial QSM cylinder graph construction from single-station TLS point clouds; (ii) multi-relation graph encoding with parent, child, and intra-branch adjacency; (iii) anchor-conditioned dual-mode decoding for continuation and branch-start generation; (iv) constrained inference with level-wise volume budgets, producing the completed QSM.}
	\label{fig:workflow}
\end{figure*}

Figure~\ref{fig:workflow} illustrates the workflow. The partial tree point cloud first undergoes leaf point removal, and the remaining woody points are used to reconstruct the QSM using TreeQSM. The input partial QSM is converted into a cylinder graph. Node features and graph adjacency matrices are passed to a skeleton-aware graph encoder. The decoder predicts, for each observed cylinder, potential missing continuation cylinders, potential new branch cylinders, branch-start probability, local missing-volume budget, and global level-wise missing volume. During inference, predicted cylinders are decoded into continuous chains and accepted under volume constraints. For paired training and evaluation, the independently fitted partial QSM undergoes an additional reference-alignment step before graph construction, as detailed in Section \ref{sec:4.1}. 

\subsection{QSM cylinder representation}

Each TreeQSM cylinder is stored using 13 raw attributes:
\begin{equation}
	\mathbf{q}_i =
	[\mathrm{id}_i, \mathrm{parent}_i, \mathrm{child}_i, r_i, \mathbf{s}_i, \mathbf{a}_i, l_i, \ell_i, b_i],
\end{equation}
where \(\mathrm{id}_i\) is the cylinder identifier, \(\mathrm{parent}_i\) and \(\mathrm{child}_i\) are topological identifiers, \(r_i\) is radius, \(\mathbf{s}_i \in \R^3\) is the start point, \(\mathbf{a}_i \in \R^3\) is the unit axis direction, \(l_i\) is length, \(\ell_i\) is branch level, and \(b_i\) is branch identifier. The endpoint is
\begin{equation}
	\mathbf{e}_i = \mathbf{s}_i + l_i \mathbf{a}_i .
\end{equation}

For learning, each cylinder is represented by an 8-dimensional geometry vector:
\begin{equation}
	\mathbf{c}_i =
	[\mathbf{s}_i, \mathbf{a}_i, l_i, r_i] \in \R^8.
\end{equation}

In addition, a 20-dimensional input feature vector is constructed for each cylinder:
\begin{equation}
	\begin{split}
		\mathbf{x}_i =
		[&\tilde{\mathbf{s}}_i,
		\tilde{\mathbf{e}}_i,
		\tilde{\mathbf{o}}_i,
		\mathbf{a}_i,
		\tilde{r}_i,
		\tilde{l}_i,
		\tilde{\ell}_i,
%		\rho_i^{root},
		\beta_i^{\mathrm{branch}}, 
		\tilde{d}_i,
		\pi_i^{\mathrm{branch}},
		\rho_i^{\mathrm{radius}},
		\tau_i^{\mathrm{turn}}],
%		\gamma_i^{TLS}
	\end{split}
\end{equation}
where \(\tilde{\mathbf{s}}_i\), \(\tilde{\mathbf{e}}_i\), and \(\tilde{\mathbf{o}}_i\) are normalized start, end, and centre coordinates; \(\tilde{r}_i\) and \(\tilde{l}_i\) are normalized radius and length; \(\tilde{\ell}_i\) is normalized branch level;  \(\beta_i^{\mathrm{branch}}\) indicates whether the cylinder starts a new branch; \(\tilde{d}_i\) is normalized topological depth; \(\pi_i^{\mathrm{branch}}\) is normalized relative position along its branch; \(\rho_i^{\mathrm{radius}}\) is the radius ratio to its parent; \(\tau_i^{\mathrm{turn}}\) is the cosine of the turning angle relative to its parent.
%\(\rho_i^{root}\) is a root indicator; and \(\gamma_i^{TLS}\) is the normalized number of TLS scan positions.

Coordinates, lengths, and radii are normalized using the centre and scale estimated from the partial QSM:
\begin{equation}
	\tilde{\mathbf{p}} = \frac{\mathbf{p}-\mathbf{c}^{\mathrm p}}{s^{\mathrm p}},
\end{equation}
where $\mathbf{p}$ denotes the original coordinate value (e.g., $\mathbf{s}_i$, $\mathbf{e}_i$, or $\mathbf{o}_i$), $\tilde{\mathbf{p}}$ is its normalized value, $\mathbf{c}^{\mathrm p}$ is the centroid of all cylinder endpoints in the partial QSM, and $s^{\mathrm p}$ is the maximum absolute coordinate offset from the centroid. The same normalization is applied to the complete QSM and missing targets to ensure that the model is invariant to tree position and scale and avoid using complete-tree information during inference.

All branch levels larger than 3 are merged into level 3:
\begin{equation}
	\tilde{\ell}_i = \min(\ell_i,3),
\end{equation}
resulting in four level classes \(\{0,1,2,3\}\).

%$$
%\ell_i^{\mathrm{cls}}=\min(\ell_i,3),
%\qquad
%\tilde\ell_i=\frac{\ell_i^{\mathrm{cls}}}{3}.
%$$

\subsection{QSM graph construction}

The partial QSM is represented as a multi-relation graph:
\begin{equation}
	\G^\mathrm p = (\V^\mathrm p, \E_\mathrm p, \E_\mathrm c, \E_\mathrm b, \X),
\end{equation}
where $\V^\mathrm p = \{v_1, v_2, \dots, v_{N}\}$ is the set of cylinder nodes ($N$ being the number of cylinders in the partial QSM), $\X \in \mathbb{R}^{N \times 20}$ is the node feature matrix with each row $\mathbf{x}_i$ corresponding to the 20-dimensional feature vector of cylinder $v_i$, and $\E_\mathrm p$, $\E_\mathrm c$, $\E_\mathrm b$ are three edge sets encoding parent-child and intra-branch relationships. Specifically, $\E_\mathrm p$ contains parent edges that connect each cylinder to its parent, $\E_\mathrm c$ contains child edges that connect each cylinder to its children, and $\E_\mathrm b$ contains intra-branch edges that connect neighbouring cylinders within the same branch.

The three edge sets are represented by adjacency matrices $\Ac_\mathrm p$, $\Ac_\mathrm c$, and $\Ac_\mathrm b$, respectively. Each adjacency matrix is row-normalized:
\begin{equation}
	\tilde{\Ac}_{ij} = \frac{\Ac_{ij}}{\max(\sum_j \Ac_{ij},1)}.
\end{equation}

\subsection{Skeleton-aware graph encoder}

The encoder first projects node features into a hidden space:
\begin{equation}
	\Hh^{(0)} = f_{\mathrm{in}}(\X),
\end{equation}
where \(f_{\mathrm{in}}\) is a two-layer multilayer perceptron (MLP) with GELU activation and \(\Hh^{(0)} \in \mathbb{R}^{N \times 128}\).

Each graph message-passing layer aggregates self, parent, child, and branch-continuity messages:
%\begin{equation}
%	\begin{split}
%		\mathbf{M}^{(t)} =
%		&\mathbf{W}_{s}\Hh^{(t)}
%		+ \mathbf{W}_{p}\tilde{\Ac}_{p}\Hh^{(t)}
%		+ \mathbf{W}_{c}\tilde{\Ac}_{c}\Hh^{(t)}  \\
%		&+ \mathbf{W}_{b}\tilde{\Ac}_{b}\Hh^{(t)},
%	\end{split}
%\end{equation}
\begin{equation}
	\begin{split}
		\mathbf M^{(t)}
		={}&
		\mathbf H^{(t)}\mathbf W_{\mathrm s}^{(t)}
		+\tilde{\mathbf A}_{\mathrm p}
		\mathbf H^{(t)}\mathbf W_{\mathrm p}^{(t)}\\
		&+\tilde{\mathbf A}_{\mathrm c}
		\mathbf H^{(t)}\mathbf W_{\mathrm c}^{(t)}
		+\tilde{\mathbf A}_{\mathrm b}
		\mathbf H^{(t)}\mathbf W_{\mathrm b}^{(t)}.
	\end{split}
\end{equation}
where \(t\) denotes the layer index and \(\Hh^{(t)} \in \mathbb{R}^{N \times 128}\) 
is the node embedding matrix at layer \(t\). $\mathbf W_{\mathrm s}^{(t)}$, $\mathbf W_{\mathrm p}^{(t)}$, $\mathbf W_{\mathrm c}^{(t)}$, and $\mathbf W_{\mathrm b}^{(t)} \in \mathbb R^{128\times128}$ are learnable, relation-specific linear transformation matrices
at message-passing layer $t$. The node embeddings are then updated 
with residual connections, layer normalization, and a feed-forward network (FFN):
\begin{equation}
	\Hh' = \mathrm{LN}\left(\Hh^{(t)} + \mathrm{Dropout}(\mathbf{M}^{(t)})\right),
\end{equation}
\begin{equation}
	\Hh^{(t+1)} =
	\mathrm{LN}\left(\Hh' + \mathrm{Dropout}(\mathrm{FFN}(\Hh'))\right),
\end{equation}
where FFN is a two-layer network with GELU activation and a hidden dimension of 512.

After \(T\) layers, the encoder outputs skeleton-aware node embeddings:
\begin{equation}
	\mathbf{Z} = \Hh^{(T)} \in \mathbb{R}^{N \times 128},
\end{equation}
which serve as the input to the decoder. In the implementation, \(T=4\) and dropout is 0.1.

\subsection{Anchor-conditioned dual-mode decoder}

The decoder predicts missing structure conditioned on each observed cylinder. Two generation modes are used.

\subsubsection{Tip continuation}

For each node \(i\), the continuation decoder predicts a chain of up to \(K_\mathrm c\) cylinders extending from the tip of an existing branch:
\begin{equation}
	\{\hat{\mathbf{c}}^{\mathrm{cont}}_{i,k}\}_{k=1}^{K_\mathrm c},
\end{equation}
where each \(\hat{\mathbf{c}}^{\mathrm{cont}}_{i,k} \in \mathbb{R}^{8}\) comprises a predicted start offset, axis, length, and radius. The start offset is ignored during inference and replaced by the hard anchoring rule (Sections \ref{sec:3.5.3} and \ref{sec:3.7}), leaving axis, length, and radius as the effective geometric parameters. For each slot \(k\), the decoder additionally predicts an existence logit \(\hat{z}^{\mathrm{cont}}_{i,k} \in \mathbb{R}\) indicating whether the \(k\)-th cylinder in the continuation chain is present and a level logit \(\hat{\mathbf{u}}^{\mathrm{cont}}_{i,k} \in \mathbb{R}^{4}\) which is used only as an auxiliary training signal and is superseded by the topological level rule during inference.

Continuation is only accepted for terminal nodes:
\begin{equation}
	m_i^{\mathrm{tip}} = \mathbb{I}(|\mathcal{C}(i)|=0),
\end{equation}
where \(\mathcal{C}(i)\) denotes the child set of node \(i\). The level of a continuation cylinder follows its parent:
\begin{equation}
	\ell_{\mathrm{new}} = \ell_{\mathrm{parent}}, \label{eq.17}
\end{equation}
ensuring that branch level remains unchanged along a continuation chain. This topology-derived level assignment prevents the classifier from introducing level inconsistencies and is used during inference to guarantee biological coherence.

\subsubsection{Branch-start generation}

The branch decoder predicts a chain of up to \(K_\mathrm b\) cylinders for a possible new branch originating from an internal node:
\begin{equation}
	\{\hat{\mathbf{c}}^{\mathrm{br}}_{i,k}\}_{k=1}^{K_\mathrm b},
\end{equation}
where each \(\hat{\mathbf{c}}^{\mathrm{br}}_{i,k} \in \mathbb{R}^{8}\) comprises a predicted start offset, axis, length, and radius. As in the continuation decoder, the start offset is ignored during inference and replaced by the hard anchoring rule. For each slot \(k\), the decoder additionally predicts an existence logit \(\hat{z}^{\mathrm{br}}_{i,k} \in \mathbb{R}\) and a level logit \(\hat{\mathbf{u}}^{\mathrm{br}}_{i,k} \in \mathbb{R}^{4}\). At the anchor level, the decoder predicts a branch-start logit \(\hat{z}^{\mathrm{start}}_{i} \in \mathbb{R}\), which indicates whether node \(i\) should initiate a new branch.

A new branch is permitted only for eligible internal nodes that have at least one child. The level of the first cylinder in a new branch is determined by a topological rule:
\begin{equation}
	\ell_{\mathrm{new}} = \min(\ell_{\mathrm{parent}} + 1, 3), \label{eq.19}
\end{equation}
ensuring that a new branch advances to the next hierarchical level. Subsequent cylinders in the chain maintain this level. As with continuation, the level logit \(\hat{\mathbf{u}}^{\mathrm{br}}_{i,k}\) is used only as an auxiliary training signal and is superseded by this topological rule during inference.

\subsubsection{Cylinder parameterization}\label{sec:3.5.3}

The decoder predicts raw cylinder parameters, which are transformed to valid geometry. The decoder does not predict the absolute start point directly. Instead, it predicts 
a start offset \(\Delta\hat{\mathbf{s}}_{i,k}\) relative to the anchor endpoint, 
which is then fixed to zero:
\begin{equation}
	\Delta\hat{\mathbf{s}}_{i,k}=\mathbf{0},
\end{equation}
because the absolute start point is determined during decoding from the endpoint of the parent cylinder. This hard anchoring ensures that all generated cylinders are physically attached to the existing structure. The axis is normalized to unit length to decouple direction from magnitude:
\begin{equation}
	\hat{\mathbf{a}}_{i,k} =
	\frac{\mathbf{a}^{\mathrm{raw}}_{i,k}}
	{\|\mathbf{a}^{\mathrm{raw}}_{i,k}\|_2+\eps},
\end{equation}
where \(\eps\) is a small positive constant (e.g., \(10^{-8}\)) added for numerical stability. The length and radius are made positive using softplus:
\begin{equation}
	\hat{l}_{i,k}=\mathrm{softplus}(l^{\mathrm{raw}}_{i,k})+\eps,
	\quad
	\hat{r}_{i,k}=\mathrm{softplus}(r^{\mathrm{raw}}_{i,k})+\eps .
\end{equation}
This parameterization ensures that all predicted cylinders are geometrically valid and physically realizable.

\subsection{Global and local missing volume prediction}
Predicting missing structure alone is insufficient for accurate volume recovery, because the model may over- or under-generate volume without global guidance. To address this, the decoder predicts two complementary volume budgets at both fine-grained and coarse-grained levels.

For each anchor node $i$, the decoder predicts a local missing-volume budget:
\begin{equation}
	\hat{\eta}_i = \mathrm{softplus}(f_{\mathrm{bud}}(\mathbf{z}_i)),
\end{equation}
where $\mathbf{z}_i \in \mathbb{R}^{128}$ is the encoded representation of anchor 
$i$ (output of the graph encoder), and $f_{\mathrm{bud}}$ is a multilayer perceptron that 
maps the anchor embedding to a scalar budget. This budget estimates how much 
missing volume should be attached to this specific anchor, and is supervised 
during training by the anchor budget loss (Section \ref{sec:3.8}).

The decoder also predicts a global volume budget for each branch level:
\begin{equation}
	\hat{\mathbf{V}}^{\mathrm{global}} =
	\mathrm{softplus}
	\left(
	f_{\mathrm{vol}}
	\left(
	\frac{\sum_i m_i \mathbf{z}_i}{\sum_i m_i}
	\right)
	\right)
	\in \mathbb{R}^{4},
\end{equation}
where $m_i$ is the node mask (1 for valid nodes, 0 for padding), and $f_{\mathrm{vol}}$ 
is a multilayer perceptron that takes the average pooled node embeddings as input 
and outputs four scalars corresponding to levels 0 through 3. This global budget 
estimates the total missing volume at each branch level for the entire tree.

The local budgets provide fine-grained supervision at the anchor level, encouraging the model to distribute volume appropriately across individual attachment points. The global budget provides a coarse-grained constraint ensuring that the total generated volume per branch level does not exceed the estimated missing volume. During inference, $\hat{\eta}_i$ limits the total accepted volume assigned to observed anchor $i$, including the generated cylinders in its associated chains. The global budget limits the combined accepted volume at each branch level across all anchors and both generation modes. A candidate is accepted only when it satisfies both the remaining anchor budget and the remaining global budget at its topology-derived branch level. The corresponding accumulated volumes are updated after acceptance.

\subsection{Missing target construction}\label{sec:3.7}

For the paired data used in this study, the partial QSM is constructed as a subset of the reference cylinders through the alignment procedure described in Section \ref{sec:4.1}. The retained cylinders share the reference identifiers. Missing cylinders are therefore identified in this shared index space, rather than by comparing the original IDs from two independent TreeQSM runs:
\begin{equation}
	\M = \{v_j \in \G^\mathrm r \mid id_j \notin \G^\mathrm p\}.
\end{equation}
Missing cylinders are sorted in topological order so that parent missing cylinders precede their children.

Each missing cylinder is assigned a parent source:
\begin{equation}
	\mathrm{source}_j \in \{0,1\},
\end{equation}
where $\mathrm{source}_j = 0$ denotes a parent in the partial QSM, and $\mathrm{source}_j = 1$ denotes a previously generated missing cylinder. For each missing chain, the anchor is obtained by tracing the parent relation back to the first observed partial cylinder. If the first missing cylinder starts a new branch, the chain is assigned to the branch decoder; otherwise, it is assigned to the continuation decoder.

The training target for generated cylinders sets the start coordinate to zero, while retaining axis, length, and radius:
\begin{equation}
	\mathbf{c}^{\mathrm{target}}_{j}
	=
	[\mathbf{0},\mathbf{a}_{j},l_j,r_j].
\end{equation}
This design enforces geometric continuity during decoding and avoids detached generated cylinders.

\subsection{Training losses}\label{sec:3.8}

The model is trained using a multi-task objective:
\begin{equation}
	\begin{split}
		\Lc =
		&\lambda_\mathrm{vol}\Lc_\mathrm{vol}
		+ \lambda_\mathrm{lvol}\left(\Lc_\mathrm{lvol}+0.5\Lc_\mathrm{gvol}\right)
		+ \lambda_\mathrm{bud}\Lc_\mathrm{bud} \\
		&+ \lambda_\mathrm{exist}\Lc_\mathrm{exist}
		+ \lambda_\mathrm{geo}\Lc_\mathrm{geo}
		+ \lambda_\mathrm{level}\Lc_\mathrm{level}
		+ \lambda_\mathrm{start}\Lc_\mathrm{start} \\
		&+ \lambda_\mathrm{smooth}\Lc_\mathrm{smooth}
		+ \lambda_\mathrm{rad}\Lc_\mathrm{rad}.
	\end{split} \label{eq:28}
\end{equation} 

All volume calculations are based on frustum volume approximation. For each cylinder with its radius as the start radius, the radius $r_\mathrm c$ of its child cylinder as the end radius, and its length $l$, the volume is computed as:
\begin{equation}
	\frac{\pi l}{3}(r^2 + r r_{\mathrm c} + r_\mathrm c^2).
\end{equation}
The volume of a generated cylinder is approximated during training as
\begin{equation}
%	V(\hat{\mathbf{c}})=\pi \hat{r}^{2}\hat{l},
	V(\hat{\mathbf{c}})=\frac{\pi \hat{l}}{3}({\hat{r}}^2 + \hat{r}\hat{r}_\mathrm c + \hat{r}_\mathrm c^2),
\end{equation}
where $\hat{r}$ and $\hat{r}_c$ are the predicted radii of itself and its child cylinder. For a terminal element with no successor in the decoded chain, the end radius is set equal to its start radius, reducing the frustum expression to a cylindrical volume. The same terminal-radius convention is used in the corresponding training and inference volume calculations. $\hat{l}$ is the predicted length of itself. Expected predicted volume is weighted by existence probability $\sigma(\hat{z}_{i,k})$ and level probability $P_{i,k}(\ell)$:
\begin{equation}
	\hat{V}^{(\ell)}
	=
	\sum_{i,k}
	V(\hat{\mathbf{c}}_{i,k})
	\sigma(\hat{z}_{i,k})
	P_{i,k}(\ell).
\end{equation}
During training, each predicted cylinder contributes to the volume at a branch level according to its existence probability and its predicted probability of belonging to that level. This allows the volume loss to provide gradients for network training. The total-volume loss is
\begin{equation}
	\Lc_\mathrm{vol}
	=
	\frac{
		|\sum_{\ell}\hat{V}^{(\ell)}
		-
		\sum_{\ell}V^{(\ell)}|
	}
	{\sum_{\ell}V^{(\ell)}+\eps}.
\end{equation}
The level-wise volume loss is
\begin{equation}
	\Lc_\mathrm{lvol}
	=
	\frac{
		\sum_{\ell=0}^{3}
		\omega_{\ell}
		|\hat{V}^{(\ell)}-V^{(\ell)}|
	}
	{\sum_{\ell}V^{(\ell)}+\eps},
\end{equation}
where $\omega_{\ell}$ is a per-level weight (set to $\omega_0=1.0$, 
$\omega_1=2.0$, $\omega_2=2.0$, $\omega_3=1.0$ to emphasize mid-level 
branches). The global volume head is supervised by
\begin{equation}
	\Lc_\mathrm{gvol}
	=
	\frac{
		\|\hat{\mathbf{V}}^\mathrm{global}-\mathbf{V}\|_1
	}
	{\sum_{\ell}V^{(\ell)}+\eps},
\end{equation}
where $\mathbf{V} = [V^{(0)}, V^{(1)}, V^{(2)}, V^{(3)}]^\top$ is the reference level-wise volume vector.

The target anchor volume is
\begin{equation}
	\eta_i = \sum_{j:a(j)=i}V_j,
\end{equation}
where \(a(j)\) is the observed anchor assigned to missing cylinder \(j\), and $V_j$ is the reference volume of missing cylinder $j$. This quantity represents the total missing volume that should be attached to anchor $i$. The budget loss is
\begin{equation}
	\Lc_\mathrm{bud}
	=
	\frac{
		\sum_{i:{\eta}_i>0}|\hat{\eta}_i-{\eta}_i|
		+
		0.1\sum_{i:{\eta}_i=0}|\hat{\eta}_i|
	}
	{\sum_{\ell}V^{(\ell)}+\eps}.
\end{equation}

Weighted binary cross-entropy is used for continuation and branch slots:
\begin{equation}
	\Lc_\mathrm{exist}
	=
	\frac{1}{|\Omega|}
	\sum_{(i,k)\in \Omega}
	\mathrm{BCE}(\hat{z}_{i,k},y_{i,k}),
\end{equation}
where $\Omega$ is the set of all anchor-slot pairs $(i,k)$, where $i$ indexes 
an anchor node and $k$ indexes a slot position within that anchor's predicted chain. For each pair $(i,k)$, the model predicts a logit $\hat{z}_{i,k}$, and the reference QSM label $y_{i,k} \in \{0,1\}$ indicates whether a cylinder exists at that position in the complete QSM. 

The geometry loss is computed only on positive slots:
\begin{equation}
	\Lc_\mathrm{geo}
	=
	\frac{1}{|\mathcal{P}|}
	\sum_{(i,k)\in\mathcal{P}}
	\|\hat{\mathbf{c}}_{i,k}-\mathbf{c}_{i,k}\|_1,
\end{equation}
where $\mathcal{P} \subset \Omega$ is the subset of anchor-slot pairs where a cylinder exists in the reference QSM (positive samples).

Cross-entropy is used for branch level classification:
\begin{equation}
	\Lc_\mathrm{level}
	=
	\frac{1}{|\mathcal{P}|}
	\sum_{(i,k)\in\mathcal{P}}
	\mathrm{CE}(\hat{\mathbf{u}}_{i,k},\ell_{i,k}).
\end{equation}

A branch-start binary cross-entropy loss supervises whether an eligible non-tip anchor should generate a new branch:
\begin{equation}
	\Lc_\mathrm{start}
	=
	\frac{1}{|\Omega_\mathrm b|}
	\sum_{i\in\Omega_\mathrm b}
	\mathrm{BCE}(\hat{z}^\mathrm{start}_{i},y^\mathrm{start}_{i}),
\end{equation}
where $\Omega_\mathrm b$ means the set of anchor nodes that are eligible for branch-start generation, i.e., non-tip nodes not belonging to forbidden branches.

For continuation cylinders, abrupt direction changes are penalized:
\begin{equation}
	\Lc_\mathrm{smooth}
	=
	\frac{1}{|\mathcal{P}_\mathrm{cont}|}
	\sum_{(i,k)\in\mathcal{P}_\mathrm{cont}}
	\left(
	1 -
	\frac{
		\hat{\mathbf{a}}_{i,k}\cdot\mathbf{a}_i
	}
	{\|\hat{\mathbf{a}}_{i,k}\|_2\|\mathbf{a}_i\|_2+\eps}
	\right).
\end{equation}

A tapering prior constrains generated radii relative to parent radius:
\begin{equation}
	\begin{split}
		\Lc_\mathrm{rad} =
		\frac{1}{|\mathcal{P}|}
		\sum_{(i,k)\in\mathcal{P}}
		&\left[\max(\alpha_\mathrm{min}r_i-\hat{r}_{i,k},0) \right. \\
		&\left. + \max(\hat{r}_{i,k}-\alpha_\mathrm{max}r_i,0) \right],
	\end{split}
\end{equation}
where $\alpha_\mathrm{min}$ and $\alpha_\mathrm{max}$ are the lower and upper radius ratio bounds, respectively. 

\subsection{Constrained inference}

During inference, the decoder output is converted into QSM cylinders using chain-based decoding. For a generated chain anchored at observed cylinder $i$, the first generated cylinder starts at the endpoint of the anchor:
\begin{equation}
	\hat{\mathbf{s}}_{i,1} = \mathbf{e}_i.
\end{equation}
Subsequent generated cylinders start at the endpoint of the previous generated cylinder:
\begin{equation}
	\hat{\mathbf{s}}_{i,k}
	=
	\hat{\mathbf{s}}_{i,k-1}
	+
	\hat{l}_{i,k-1}\hat{\mathbf{a}}_{i,k-1}.
\end{equation}

During inference, a cylinder is either accepted or rejected, and its branch level is determined from its parent–child relationship using Equations. (\ref{eq.17}) and (\ref{eq.19}). The resulting branch-level volumes can therefore differ from the probability-weighted volumes used during training. The training losses encourage agreement with target volumes, while the inference checks separately limit accepted volume relative to the predicted budgets. Candidate cylinders are accepted according to existence probability, branch-start probability, topological eligibility, and level-wise volume budgets. A candidate cylinder at level \(\ell\) is accepted only if the accumulated generated volume remains within the corresponding budget. Algorithm~\ref{alg:inference} summarizes the inference procedure. The budgets are upper bounds rather than exact volume-filling targets. A candidate that would exceed a remaining budget is not accepted. The final accepted cylinder is not resized to consume the residual budget, and unused budget is allowed at termination.
%The predicted global level-wise volume \(\hat{\mathbf{V}}^{global}\) is scaled and split into continuation and branch-start budgets:
%\begin{equation}
%	B_{\ell}^{cont} = \gamma_{\ell}^{cont}\hat{V}_{\ell}^{global},
%	\quad
%	B_{\ell}^{br} = \gamma_{\ell}^{br}\hat{V}_{\ell}^{global}.
%\end{equation}

%Additional structural constraints are applied:
%\begin{itemize}
%	\item continuation cylinders are constrained to lie within a cone around the parent axis;
%	\item the first cylinder of a new branch must form a minimum and maximum branching angle relative to the parent axis;
%	\item subsequent cylinders in a generated branch are constrained to be smooth;
%	\item generated radii are clamped according to tapering ratios;
%	\item generated lengths are capped relative to parent length;
%	\item specified branches, typically the trunk, can be prevented from receiving new side branches.
%\end{itemize}

\begin{algorithm}[htbp]
	\caption{Volume-aware anchor-conditioned QSM completion}
	\label{alg:inference}
	\begin{algorithmic}[1]
%		\Require Partial QSM \(\G^p\), trained model \(f_\theta\), maximum rounds \(R\)
		\Require Partial QSM \(\G^\mathrm p\), trained model \(f_\theta\)
		\State \(\hat{\G}^\mathrm c\leftarrow \G^\mathrm p\)
%		\For{\(r=1,\ldots,R\)}
		\State Build node features and graph adjacency for \(\hat{\G}^\mathrm c\);
		\State Predict continuation cylinders, branch cylinders, existence scores, branch-start scores, and level-wise volume budget using \(f_\theta\);
		\State Identify tip anchors and eligible branch-start anchors;
		\State Generate candidate continuation and branch chains;
%		\State Apply angle, length, radius, branch-level, and attachment constraints
		\State Rank candidate cylinders in descending order of their predicted existence probabilities;
		\State Accept candidates subject to level-wise continuation and branch-start volume budgets;
		\State Merge accepted cylinders into \(\hat{\G}^\mathrm c\);
		\If{no cylinder is accepted};
		\State break
		\EndIf
%		\EndFor
		\State \Return completed QSM \(\hat{\G}^\mathrm c\)
	\end{algorithmic}
\end{algorithm}

%\subsection{QSM volume computation}

%During training, differentiable cylinder volume is used:
%\begin{equation}
%	V_{cyl} = \pi r^2 l.
%\end{equation}
%For final QSM reporting and mesh export, a frustum volume can be computed:
%\begin{equation}
%	V_{frustum}
%	=
%	\frac{\pi l}{3}
%	(r_0^2+r_0r_1+r_1^2),
%\end{equation}
%where \(r_0\) is the start radius and \(r_1\) is the resolved end radius from the child cylinder or a tapered tip radius.

% ============================================================
\section{Experimental setup} \label{sec:4}
% ============================================================
% ============================================================
\subsection{Datasets} \label{sec:4.1}
% ============================================================
The proposed method is validated using real tree point clouds and simulated data.

\subsubsection{Real TLS urban tree dataset} \label{sec:4.1.1}
\begin{figure*}[!ht]
	\centering
	\includegraphics[width=1\linewidth]{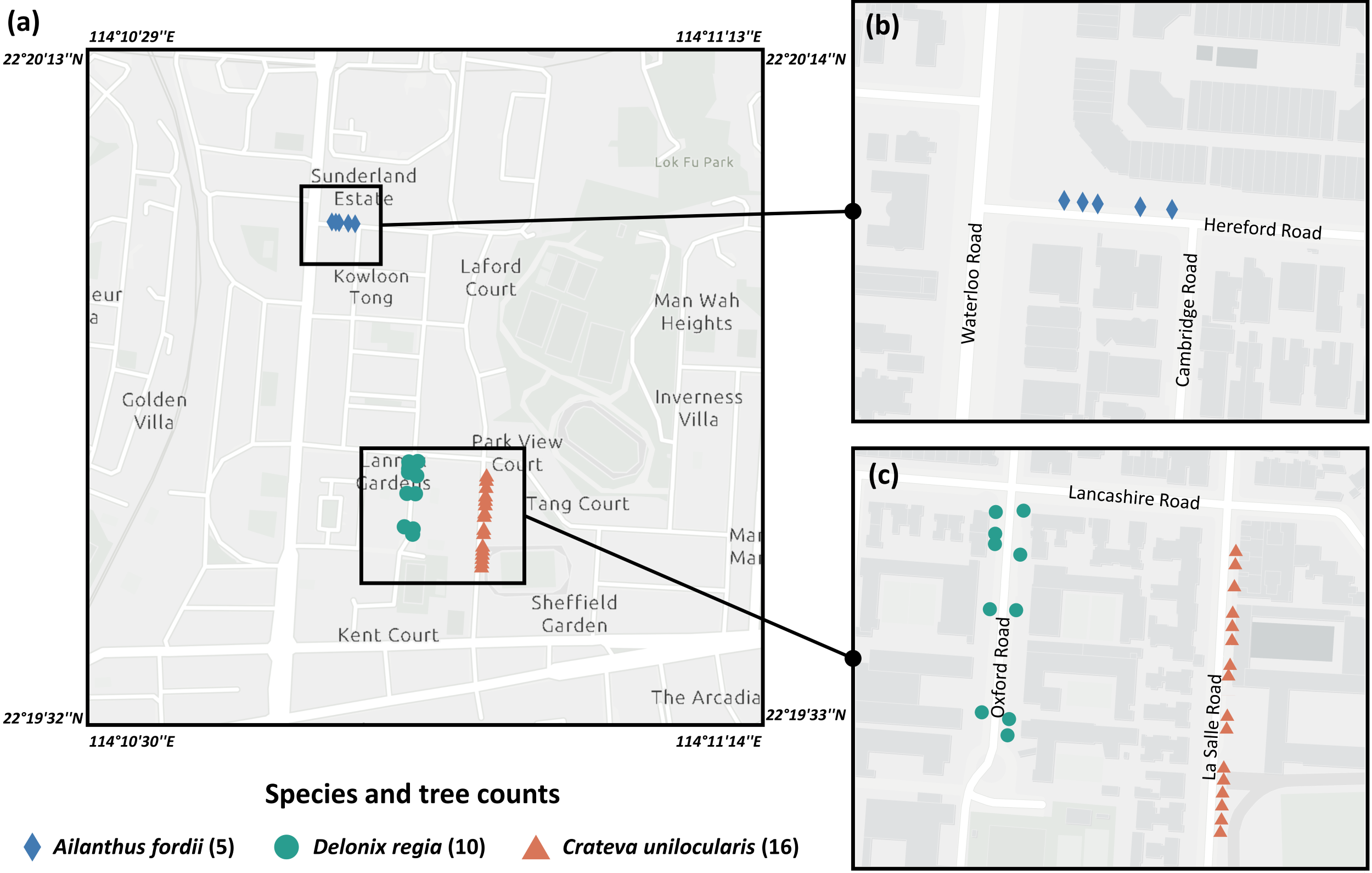}
	\caption{Locations of the urban trees collected in Kowloon Tong, Hong Kong. (a) Overview map showing the approximate locations of the three target tree species; (b) Enlarged map showing the individual locations of 5 \textit{Ailanthus fordii}; (c) Enlarged map showing the individual locations of 10 \textit{Delonix regia} and 16 \textit{Crateva unilocularis}.}
	\label{fig:Map_all}
\end{figure*}

The real dataset consists of TLS scans of urban trees collected in February 2025, Kowloon Tong, Hong Kong. Three target tree species were scanned, including \textit{Ailanthus fordii}, \textit{Delonix regia}, and \textit{Crateva unilocularis}. Figure~\ref{fig:Map_all} shows the locations of the collected urban trees in Kowloon Tong, Hong Kong, including an overview of the sampling sites and detailed maps of the individual tree locations. Tree point clouds were acquired with a Leica RTC360 terrestrial laser scanner. Scans used the high-resolution field mode, with a nominal point spacing of approximately 3 mm at 10 m and an angular sampling interval of 0.0172 degrees. Each campaign used at least three to four same-day viewpoints around the accessible sides of a target tree row. For each tree, multiple TLS stations were registered into a common coordinate system. Wood/leaf separation \citep{hu2020randla} was performed to obtain woody point clouds before QSM reconstruction. A QSM reconstructed from one TLS station was regarded as a partial observation, while the QSM reconstructed from all registered stations was regarded as the reference QSM. 

Let \(\mathbf{C}_{\tau}^{(s)}\) denote the woody point cloud of tree \(\tau\) from station \(s\), and let \(\mathbf{C}_{\tau}^{\mathrm{all}}=\bigcup_s \mathbf{C}_{\tau}^{(s)}\) denote the multi-station point cloud. TreeQSM was applied to both \(\mathbf{C}_{\tau}^{(s)}\) and \(\mathbf{C}_{\tau}^\mathrm{all}\), yielding paired partial and reference QSMs:
\begin{equation}
	\tilde{\G}_{\tau}^\mathrm p = \mathrm{TreeQSM}(\mathbf{C}_{\tau}^{(s)}).
\end{equation}

\begin{equation}
	\G_{\tau}^\mathrm c = \mathrm{TreeQSM}(\mathbf{C}_{\tau}^\mathrm{all}).
\end{equation}
The independently fitted partial QSM, $\tilde{\G}_{\tau}^\mathrm p$, is subsequently aligned with the reference QSM to construct the final paired input. Specifically, reference cylinders corresponding to the observed regions represented by $\tilde{\G}_{\tau}^\mathrm p$ are selected to form $\G_{\tau}^\mathrm c$. The selected cylinders retain their reference geometry and identifiers. Thus, the initial partial QSM identifies the observed structure, whereas the final partial QSM expresses that structure in the reference cylinder representation.

Dataset splitting was performed by tree identity. All station-specific point clouds, partial QSMs, and multi-station references associated with the same tree were assigned to the same subset. Consequently, no observation derived from a test tree was included in the training or validation set.

\subsubsection{Simulated TLS dataset from DART}

The simulated dataset was generated using the DART \citep{gastellu2016simulation, yin2016simulation, yin2019modeling} radiative transfer model. The object files of four tree models, i.e., \textit{Acer palmatum}, \textit{Betula pendula}, \textit{Populus nigra}, and \textit{Populus tremuloides}, were used as input with detailed DART scene configurations shown in Table \ref{tab:DART_scene_configuration}. Similar to the real dataset, a single simulated TLS station was used to generate a partial tree point cloud with post-processing to remove leaf points; however, a complete tree point cloud is generated by directly sampling points from the integral surface of the tree obj model without leaves. The sampling density of complete point cloud is consistent with the simulated point cloud density of partial point cloud. TreeQSM was then applied to obtain paired partial and complete QSMs. The same correspondence-based construction described in Section \ref{sec:4.1.1} is applied to the independently reconstructed simulated QSMs to obtain reference-aligned partial inputs with shared cylinder identifiers. The simulated dataset contained 26 geometrically distinct source meshes rather than repeated scan realizations of a single mesh. Each source mesh represented an independent tree individual. All partial scans and point-sampling realizations derived from one source mesh were assigned to the same data subset, ensuring that no geometry from a test source mesh was used during training or validation.

\begin{table}[H]
	\centering
	\caption{Configuration parameters of the DART radiative transfer model for simulating TLS point clouds. A single position was used to generate the partial observation.}
	\label{tab:DART_scene_configuration}
	\begin{tabular}{lc}
		\toprule
		Configuration & Description \\
		\midrule
		Acquisition rate (period) & 0.02 ns\\
		Pulse relative power  & 0.5\\
		Pulse duration at relative power & 0.05 ns \\
		Spectral bandwidth & 1.55 $\mu$m \\
		LiDAR position (x, y, z) & (10 m, 20 m, 1 m) \\
		Resolution zenith & $0.02^\circ$\\
		Resolution azimuth & $0.02^\circ$\\
		Number of scan positions & 1 \\
		Central azimuth (target to TLS) & ($0^\circ$, $90^\circ$, $120^\circ$, $180^\circ$) \\
		Delta azimuth & $90^\circ$ \\
		Central zenith (target to TLS) & $100^\circ$ \\
		Delta zenith & $120^\circ$ \\
		Divergence & $0.01^\circ$ \\
		Field of view & $0.02^\circ$ \\
		\bottomrule
	\end{tabular}
\end{table}

%\subsubsection{Dataset split}

Both real and simulated datasets were divided into training, validation, and test sets. The number of corresponding trees are listed in Table \ref{tab:dataset_summary}.
%\begin{table}[htbp]
%	\centering
%	\caption{Summary of datasets used in this study.}
%	\label{tab:dataset_summary}
%	\begin{tabular}{llcccc}
%		\toprule
%		Dataset & Source & Trees & Train & Val. & Test \\
%		\midrule
%		Sim-TLS & DART simulation & 26 & 18 & 4 & 4  \\
%		Real-TLS & Hong Kong TLS & 31 & 19 & 6 & 6 \\
%		\bottomrule
%	\end{tabular}
%\end{table}

\begin{table*}[H]
	\centering
	\caption{Summary of datasets used in this study.}
	\label{tab:dataset_summary}
	\begin{tabular}{lllccccc}
		\toprule
		Dataset & Source & Species name & Abbr. & Qty. & Train & Val. & Test \\
		\midrule
		\multirow{4}{*}{Sim-TLS} &
		\multirow{4}{*}{DART simulation} 
		& \textit{Acer palmatum}  & $\mathrm{ST}_1$ & 7 & 5 & 1 & 1 \\
		& & \textit{Betula pendula} & $\mathrm{ST}_2$ & 7  & 5 & 1 & 1 \\
		& & \textit{Populus nigra} & $\mathrm{ST}_3$ &  7 & 5 & 1 & 1 \\
		& & \textit{Populus tremuloides} & $\mathrm{ST}_4$ & 5  & 3 & 1 & 1 \\
		\midrule
		\multirow{3}{*}{Real-TLS}  &
		\multirow{3}{*}{Hong Kong TLS} 
		& \textit{Ailanthus fordii} & $\mathrm{RT}_1$ &  5 & 3 & 1 & 1 \\
		& & \textit{Delonix regia} & $\mathrm{RT}_2$ &  10 & 6 & 2 & 2 \\
		& & \textit{Crateva unilocularis} & $\mathrm{RT}_3$ &  16 & 10 & 3 & 3\\
		\bottomrule
	\end{tabular}
\end{table*}

\subsection{Implementation details}
\begin{table}[H]
	\centering
	\caption{Main model and training hyperparameters.}
	\label{tab:hyperparameters}
	\begin{tabular}{ll}
		\toprule
		Parameter & Value \\
		\midrule
		Input feature dimension & 20 \\
		Hidden dimension & 128 \\
		Graph layers & 4 \\
		Dropout & 0.1 \\
		Continuation slots \(K_c\) & 5 \\
		Branch slots \(K_b\) & 20 \\
		Branch level classes & 4 \\
		Optimizer & AdamW \\
		Learning rate & \(1\times10^{-4}\) \\
		Weight decay & \(1\times10^{-4}\) \\
		Batch size & 10 \\
		Epochs & 1000 \\
		Seed & 42 \\
		\bottomrule
	\end{tabular}
\end{table}

The model was implemented in PyTorch. The hidden dimension was 128, the graph encoder contained four message-passing layers, and dropout was set to 0.1. The model was trained using AdamW with learning rate \(1\times10^{-4}\), weight decay \(1\times10^{-4}\), batch size 10, and 1000 epochs. Training was parallelized across two NVIDIA RTX A6000 GPUs (48 GB memory each) via PyTorch's DataParallel with per-GPU memory consumption approximately 32 GB. The checkpoint with the lowest validation loss was selected for testing. The random seed was set to 42. Table \ref{tab:hyperparameters} lists the main model and training settings. We used the following loss weights:
\[
\lambda_\mathrm{vol}=3.0,\quad
\lambda_\mathrm{lvol}=1.5,\quad
\lambda_\mathrm{bud}=0.1,
\]
\[
\lambda_\mathrm{exist}=2.0,\quad
\lambda_\mathrm{geo}=3.0,\quad
\lambda_\mathrm{level}=0.5,
\]
\[
\lambda_\mathrm{start}=2.0,\quad
\lambda_\mathrm{smooth}=1.0,\quad
\lambda_\mathrm{rad}=1.0.
\]

The framework is computationally efficient and lightweight. The model contains 905,035 trainable parameters. Network inference time should be distinguished from the runtime of point-cloud preprocessing, TreeQSM reconstruction, graph construction, and constrained decoding. On our hardware configuration, the average network inference time for a single tree is only a few milliseconds, demonstrating its potential for real-time applications.

%\subsection{Occlusion-aware data augmentation}
%
%During training, the partial QSM was augmented to simulate TLS incompleteness. Three degradation operators were used:
%\begin{itemize}
%	\item branch dropping: removing an entire non-trunk branch and its descendants;
%	\item subtree dropping: removing a subtree rooted at a non-trunk node;
%	\item tail cutting: removing short terminal segments from branch tips.
%\end{itemize}
%A minimum keep ratio was enforced to prevent excessively degraded samples:
%\begin{equation}
%	\frac{|\V^{aug}|}{|\V^{p}|} \geq 0.45.
%\end{equation}
%No augmentation was applied to validation or test samples.

\subsection{Baseline methods}

The proposed method is compared with QSM, regression-based, and point cloud-based baselines:
\begin{enumerate}
	\item Partial QSM: the input partial QSM without completion.
	\item Regression-based baseline: a traditional forestry approach that scales the total volume of the partial QSM based on a combined variable allometric model. 
	\item Point completion + TreeQSM: point cloud completion using PoinTr-based \citet{bornand2024completing}, AdaPoinTr-based \citet{10232862}, and SnowflakeNet-based \citet{xindi2025advanced} methods, followed by TreeQSM reconstruction.
	\item Proposed QSM completion: the proposed volume and skeleton-aware QSM completion method.
%	\item Complete QSM: the multi-station TreeQSM reference.
\end{enumerate}

The regression-based baseline leverages a log-linear regression utilizing DBH and tree height, serving as a non-geometric volume correction benchmark. The empirical coefficients are pre-fitted using a separate validation subset of fully intact multi-station reference trees to ensure a robust and unbiased estimation of the unobserved structures. For the point cloud completion baselines, we include PoinTr \citep{yu2021pointr}, AdaPoinTr \citep{10232862}, and SnowflakeNet \citep{xiang2021snowflakenet}, following the baseline configurations established in \citet{bornand2024completing} and \citet{bornand2026reconstructing}. As the enhanced SnowflakeNet \citep{xiang2021snowflakenet} variant proposed by \citet{xindi2025advanced} is not publicly available, we evaluate the original SnowflakeNet architecture. For each point cloud completion method, we apply the completed point cloud to TreeQSM reconstruction to obtain cylinder-based QSM models, which are then evaluated using the same metrics as our method.

\subsection{Evaluation Metrics}

To comprehensively evaluate the completion performance of our proposed method, we adopt five complementary metrics that capture different aspects of reconstruction quality. These metrics are deliberately designed to be distribution-level rather than instance-level, reflecting a fundamental characteristic of the tree completion problem, namely that given a partially observed tree structure, there exists a multiplicity of valid complete structures, and the reference QSM represents only one plausible realization. Consequently, our evaluation does not require one-to-one correspondence between individual cylinders or branches in the completed model and the reference QSM.

Let $\mathcal{G}^\mathrm p$ denote the input partial tree cylinder graph, $\mathcal{G}^\mathrm c$ denote the complete reference cylinder graph, and $\hat{\mathcal{G}}$ denote the completed graph produced by our method. Throughout this study, the term ``reference QSM” denotes the QSM reconstructed from the complete simulated point cloud or the registered multi-station TLS point cloud. For the real trees, this reference is not a direct measurement of physical woody volume and remains subject to point-cloud acquisition errors, TreeQSM parameter sensitivity, and known QSM uncertainty, particularly for small branches. We define $V(\mathcal{G})$ as the total volume of all cylinders in graph $\mathcal{G}$, and $V_{\ell}(\mathcal{G})$ as the total volume contributed by cylinders at hierarchical level $\ell \in \{0,1,2,3\}$. 

\subsubsection{Total Volume Relative Error}
The total volume relative error measures the accuracy of the global volumetric recovery:
\begin{equation}
	\mathrm{E_V} = \frac{\left|V(\hat{\mathcal{G}}) - V(\mathcal{G}^\mathrm c)\right|}{V(\mathcal{G}^\mathrm c) + \eps} \times 100\%.
\end{equation}
This is a critical metric for forestry applications, as total volume is the primary determinant of above-ground biomass estimation. An $\mathrm{E_V}$ close to zero indicates that the completed model preserves the overall volumetric integrity of the tree. The mean $\mathrm{E_V}$ value of multiple trees is referred to m$\mathrm{E_V}$.

\subsubsection{Level-wise Volume Error}
A central contribution of our method is the level-wise volume budgeting mechanism, which enforces accurate volume distribution across hierarchical branch orders. To evaluate this, we compute the mean absolute relative error across four hierarchical levels:
\begin{equation}
	\mathrm{E_{LV}} = \frac{1}{4}\sum_{\ell=0}^{3} \frac{\left|V_{\ell}(\hat{\mathcal{G}}) - V_{\ell}(\mathcal{G}^\mathrm c)\right|}{V_{\ell}(\mathcal{G}^\mathrm c) + \eps} \times 100\%.
\end{equation}
This metric captures whether the model correctly distributes volume across different branch orders, which is an aspect that total volume alone cannot reveal. A low $\mathrm{E_{LV}}$ indicates that the completed tree preserves both the total volume and its hierarchical organization, which is essential for biologically plausible reconstructions. The mean $\mathrm{E_{LV}}$ value of multiple trees is represented as m$\mathrm{E_{LV}}$.

\subsubsection{Volume Relative Bias}
The volume relative bias measures the signed relative deviation of the predicted total volume from the reference QSM:
\begin{equation}
	\mathrm{B_V} = \frac{V(\hat{\mathcal{G}}) - V(\mathcal{G}^\mathrm c)}{V(\mathcal{G}^\mathrm c) + \eps} \times 100\%,
\end{equation}
where $\mathrm{B_V} > 0$ indicates over-generation, $\mathrm{B_V} < 0$ indicates under-generation, and $\mathrm{B_V} = 0$ indicates perfect agreement. Unlike the absolute relative error $\mathrm{E_V}$, $\mathrm{B_V}$ preserves the direction of the deviation and thus characterizes whether the model systematically overestimates or underestimates woody volume. The mean $\mathrm{B_V}$ across multiple trees is denoted as m$\mathrm{B_V}$.

\subsubsection{Missing Volume Recovery.}
The missing volume recovery quantifies how much of the originally missing volume is recovered:
\begin{equation}
	\mathrm{MVR} = 
	\begin{cases}
		\dfrac{V(\hat{\mathcal{G}}) - V(\mathcal{G}^\mathrm p)}{V(\mathcal{G}^\mathrm c) - V(\mathcal{G}^\mathrm p) + \eps} \times 100\%, & \text{if } V(\hat{\mathcal{G}}) \ge V(\mathcal{G}^\mathrm p) \\[2ex]
		0, & \text{if } V(\hat{\mathcal{G}}) < V(\mathcal{G}^\mathrm p)
	\end{cases}
\end{equation}
A value near 100\% indicates near-perfect recovery of the missing volume. When the missing volume in the partial input is very small, the denominator becomes small and MVR can exceed 100\% due to numerical sensitivity. Because per-tree MVR is unstable when the missing-volume denominator is small, we additionally report aggregate missing-volume recovery:
\begin{equation}
	\mathrm{MVR_{agg}} = 
		\dfrac{\sum_i (V(\hat{\mathcal{G}_i}) - V(\mathcal{G}_i^{\mathrm p}))}{\sum_i (V(\mathcal{G}_i^{\mathrm c}) - V(\mathcal{G}_i^{\mathrm p})) + \eps} \times 100\%
\end{equation}
where $V(\hat{\mathcal{G}_i})$ represents the completed QSM volume of the $i$-th tree, and $V(\mathcal{G}_i^{\mathrm p})$ and $V(\mathcal{G}_i^{\mathrm c})$ are the corresponding partial input and complete reference volumes, respectively.

\subsubsection{Skeleton Geometry Distance}
To evaluate the geometric fidelity of the completed structure beyond volumetric measures, we compute the Chamfer distance between the skeleton point sets extracted from the completed model and the reference QSM. To reduce sensitivity to the number and length of TreeQSM cylinders, the QSM centreline is uniformly resampled at a fixed spatial interval of 0.1 m. Let $S_{\mathcal{G}}$ denote the resulting set of centreline samples for graph $\mathcal{G}$. The skeleton distance is computed between the uniformly sampled completed and reference centrelines. The Chamfer distance is:
\begin{equation}
\begin{split}
	\mathrm{CD_{ske}}(S_{\hat{\mathcal{G}}}, S_{\mathcal{G}^\mathrm c}) = 
	\frac{1}{|S_{\hat{\mathcal{G}}}|}
	\sum_{\mathbf{\hat{p}} \in S_{\hat{\mathcal{G}}}}
	\min_{\mathbf{p} \in S_{\mathcal{G}^\mathrm c}}
	\|\mathbf{\hat{p}} - \mathbf{p}\|_2 \\
	+ 
	\frac{1}{|S_{\mathcal{G}^\mathrm c}|}
	\sum_{\mathbf{p} \in S_{\mathcal{G}^\mathrm c}}
	\min_{\mathbf{\hat{p}} \in S_{\hat{\mathcal{G}}}}
	\|\mathbf{\hat{p}} - \mathbf{p}\|_2,
\end{split}
\end{equation}
where $\mathbf{\hat{p}}$ and $\mathbf{p}$ are the centreline samples for completed and reference graph $\mathcal{G}$. This metric evaluates how closely the overall geometry of the completed skeleton matches the reference QSM without requiring explicit correspondence between individual cylinders. It captures structural deviations such as misplaced branches or geometrically distorted structures that may not be reflected in volumetric metrics alone. The metric is computed in physical coordinates and is reported in metres. m$\mathrm{CD_{ske}}$ is calculated by the mean $\mathrm{CD_{ske}}$ value of multiple trees.

\subsubsection{Discussion on Metric Selection}
The evaluation does not use cylinder-count errors or direct per-cylinder comparisons, because independent QSM reconstructions may represent the same woody structure using different cylinder subdivisions. Instead, we report total and branch-order volumes together with a uniformly resampled centreline distance. These metrics reduce dependence on explicit cylinder correspondence, but they do not provide a comprehensive assessment of branching topology or structural fidelity. In particular, branching-angle distributions, taper profiles, and height-wise or voxel-wise volume distributions are not directly evaluated in this study.

% ============================================================
\section{Results} \label{sec:5}
% ============================================================

\subsection{Completion performance on simulated TLS trees and real Hong Kong TLS trees}
\begin{figure*}[htbp]
	\centering
	\includegraphics[width=0.8\linewidth]{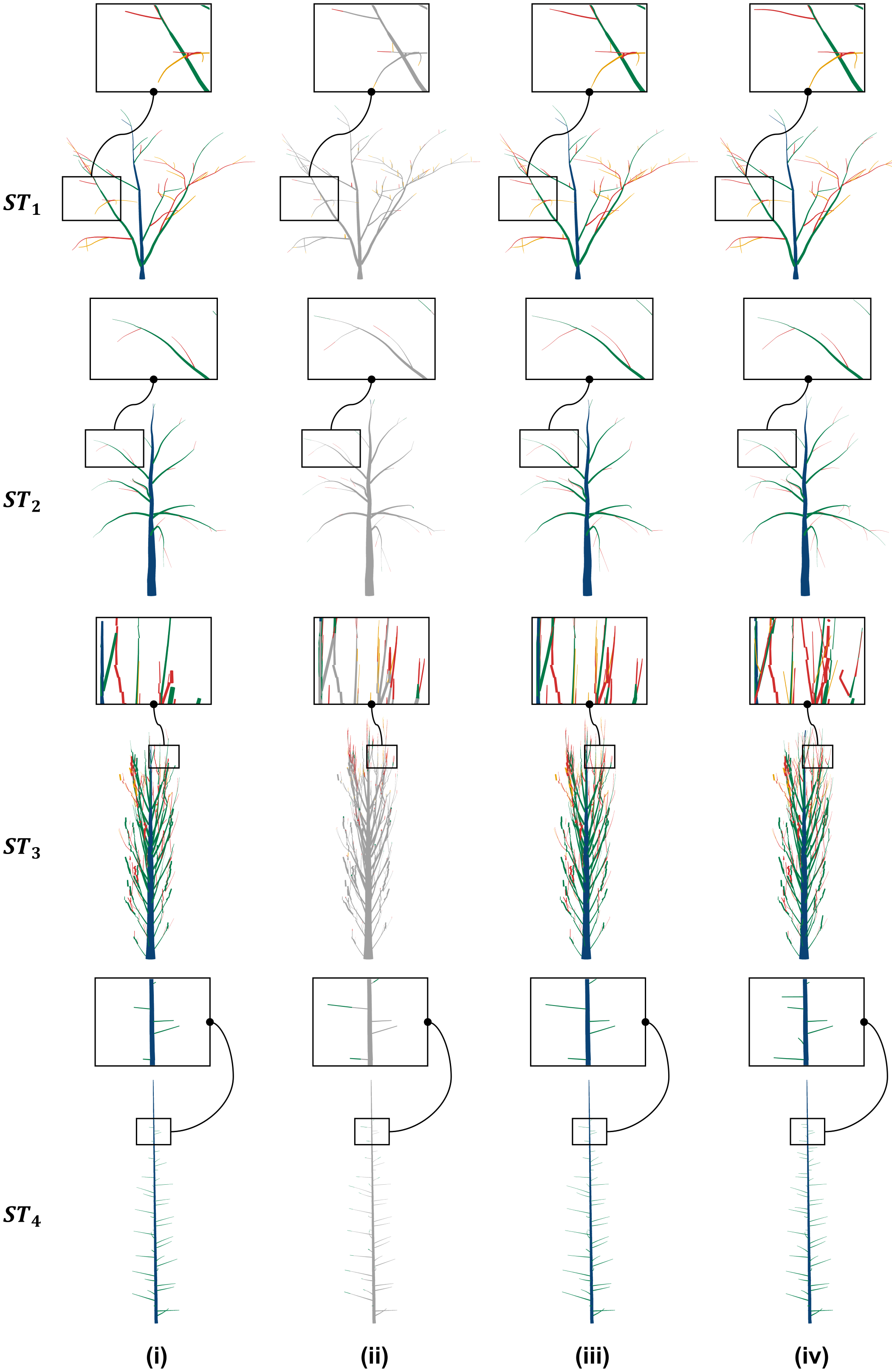}
	\caption{Qualitative comparison of tree completion results on the simulated DART dataset. $\mathrm{ST}_1$, $\mathrm{ST}_2$, $\mathrm{ST}_3$, and $\mathrm{ST}_4$ represent \textit{Acer palmatum}, \textit{Betula pendula}, \textit{Populus nigra}, and \textit{Populus tremuloides}, respectively. For each tree, we show: (i) partial QSM from single-station TLS, (ii) predicted QSM completion result with generated cylinders highlighted and original partial QSM in gray, (iii) predicted QSM completion result colored by level, and (iv) reference QSM reconstructed from points sampled from the leaf-free source mesh. Colors indicate branch levels: blue = level 0 (trunk), green = level 1, red = level 2, orange = level 3. Selected details are marked with bold outlines.}
	\label{fig:Results_sim}
\end{figure*}

\begin{figure*}[!ht]
	\centering
	\includegraphics[width=0.85\linewidth]{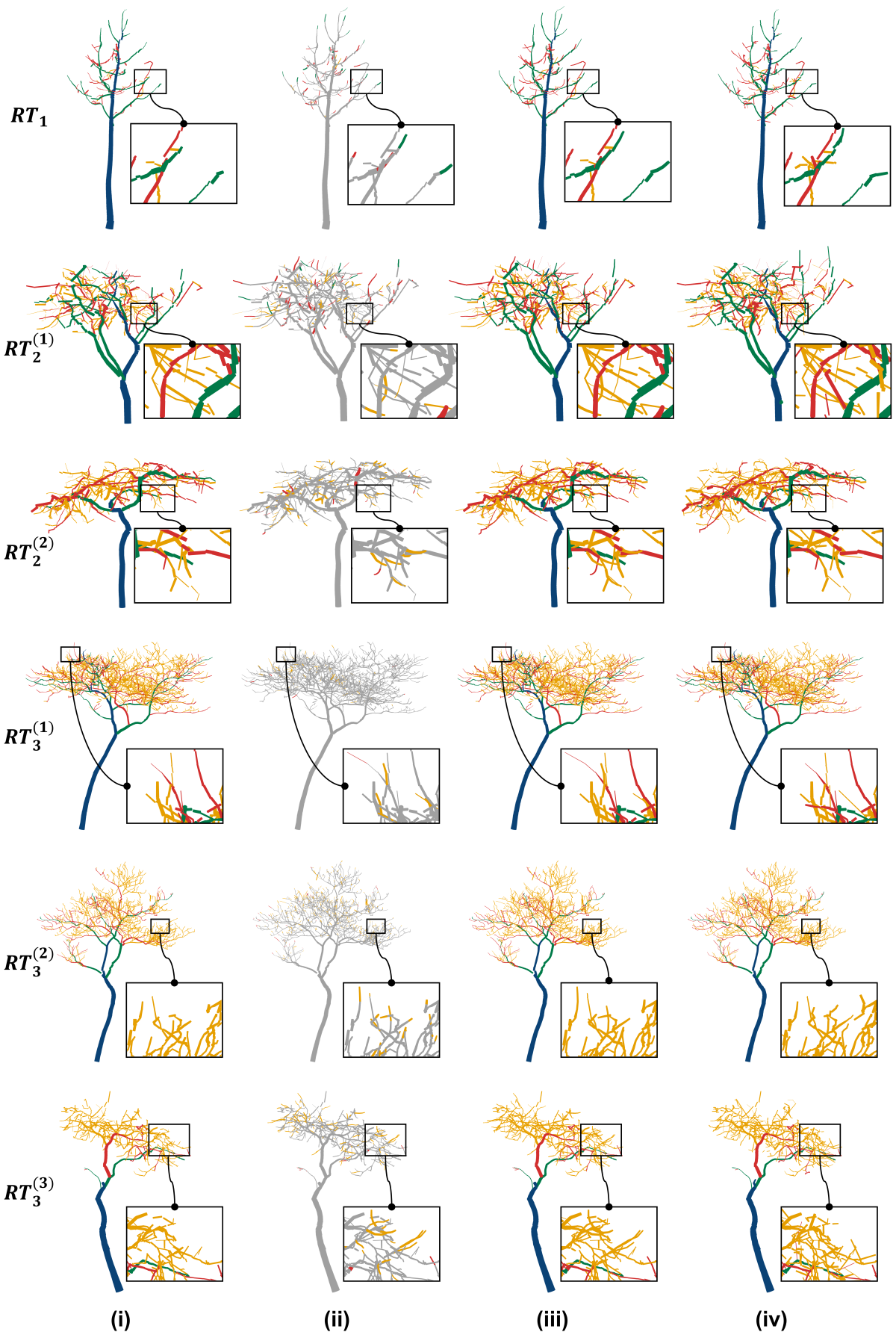}
	\caption{Qualitative comparison of tree completion results on the real Hong Kong TLS dataset. $\mathrm{RT}_1$, $\mathrm{RT}_2$, and $\mathrm{RT}_3$ represent \textit{Ailanthus fordii}, \textit{Delonix regia}, and \textit{Crateva unilocularis}, respectively. For each tree, we show: (i) partial QSM from single-station TLS, (ii) predicted QSM completion result with generated cylinders highlighted and original partial QSM in gray, (iii) predicted QSM completion result colored by level, and (iv) reference multi-station QSM. Colors indicate branch levels: blue = level 0 (trunk), green = level 1, red = level 2, orange = level 3. Selected details are marked with bold outlines.}
	\label{fig:Results_real}
\end{figure*}

The completion results for the DART simulated dataset and real TLS trees are shown in Figure~\ref{fig:Results_sim} and Figure~\ref{fig:Results_real}. Generated cylinders are highlighted using a distinct colour. A branch-level colour map is used, with level 0 for trunk, level 1 for primary branches, level 2 for secondary branches, and level 3 for high-order branches. Table~\ref{tab:volume info} reports partial, completed, and reference volumes at each branch level and for the whole tree, together with partial-to-reference and completed-to-reference volume ratios. Table \ref{tab:Performance metrics} reports the corresponding per-tree evaluation metrics. 

\subsubsection{Volume recovery analysis}

Table~\ref{tab:volume info} reports the per-level and total volumes for partial inputs, predicted outputs, and complete references, along with the corresponding recovery ratios. Several observations can be made from these results.

First, the proposed method achieves consistently high total volume recovery across all tree species. The Completed/Reference volume ratios range from 99.509\% to 100.240\%, with a mean value of 99.970\%, corresponding to a mean volume relative bias $\mathrm{B_V}$ of $-0.030$\%. This indicates that the predicted QSM preserves the total woody volume with negligible bias. In contrast, the partial inputs exhibit lower ratios, ranging from 91.906\% to 99.826\%, with a mean value of 96.584\%. The improvement in volume recovery is particularly pronounced for trees with branch-level incompleteness, such as \textit{Delonix regia} ($\mathrm{RT}_2^{(2)}$), where the partial input captures only 91.906\% of the total volume while the completed output reaches 100.010\%.

Second, the level-wise volume distribution is well preserved, especially for lower branch levels. For level 0 (trunk), the Completed/Reference ratio exceeds 99.9\% in all cases, often reaching 100\% when the trunk is already fully observed in the partial input. The near-100\% level-0 ratios primarily reflect the construction and retention of the aligned observed input. When all reference trunk cylinders are selected during alignment, the partial/reference ratio is 100\% by construction. These entries therefore do not measure independent single-station trunk-fitting accuracy or the reconstruction of newly generated trunk cylinders. For level 1 branches, the recovery ratio remains high (96.613\%--100.018\%), confirming that the model can effectively recover primary branches from partial observations.

Third, the model demonstrates its ability to recover higher-order branches (levels 2 and 3), which are the most challenging due to occlusion. For level 2, the ratio of Completed/Reference ranges from 98.972\% to 101.910\%, with a mean value of 100.039\%. For level 3, despite the inherent difficulty of recovering fine twigs, the method achieves ratios from 100.002\% to 109.498\%, with a mean value of 101.772\%. For \textit{Acer palmatum}, the partial QSM contained 91.407\% of the reference level-3 volume, whereas the completed QSM reached 109.498\%. Thus, the method overestimated the reference level-3 volume by approximately 9.5\% in this case, despite maintaining a small total-volume error. This example illustrates why total-volume agreement alone does not guarantee accurate level-wise structural recovery.

%Fourth, the missing volume recovery ratio ($R_V^{miss}$), shown in Table~\ref{tab:Performance metrics}, further confirms the effectiveness of the volume budget mechanism. For trees with substantial missing volume (e.g., \textit{Delonix regia} ($RT_2^1$) with partial/complete ratio of 92.4558\%), the model recovers 99.5627\% of the missing volume. For trees where the partial input already captures most of the structure (e.g., \textit{Betula pendula} ($ST_2$) with partial/complete ratio of 99.8263\%), the model correctly generates minimal additional volume ($R_V^{miss} = 100.1679\%$), indicating near-perfect recovery without significant over-generation. For trees where the partial input is already complete (e.g., \textit{Crateva unilocularis} ($RT_3^2$) with partial/complete ratio of 96.8870\%), the model achieves $R_V^{miss} = 100.0288\%$, correctly indicating that no additional volume is needed.

The predicted budget limits accepted volume at each level but does not guarantee complete budget utilization or eliminate error relative to the reference missing volume. Under-generation may occur when candidate confidence is low or when the remaining budget is smaller than the next candidate cylinder.

\begin{table*}[H]
	\centering
	\caption{The partial, completed, and reference volumes at each branch level and the total level, together with the ratios of partial and completed volumes to the reference volume.}
	\label{tab:volume info}
%	\begin{tabularx}{lcccccc}
\begin{tabular}{l >{\centering\arraybackslash}p{0.6cm} >{\centering\arraybackslash}p{1.4cm} >{\centering\arraybackslash}p{1.4cm} >{\centering\arraybackslash}p{1.4cm} >{\centering\arraybackslash}p{2.7cm} >{\centering\arraybackslash}p{2.7cm}}
	\toprule
	Tree species & Level & Partial ($\mathrm{m^3}$) & Completed ($\mathrm{m^3}$) & Reference ($\mathrm{m^3}$) & Partial/Reference (\%) & Completed/Reference (\%) \\
	\midrule
	& 0 & 0.127 & 0.127 & 0.127 & 99.926 & 99.942  \\
	& 1 & 0.146 & 0.146 & 0.146 & 99.788 & 99.901  \\
	\textit{Acer palmatum} ($\mathrm{ST}_1$) & 2 & 0.043 & 0.044 & 0.045 & 96.424 & 98.972  \\
	& 3 & 0.014 & 0.017 & 0.016 & 91.407 & 109.498 \\
	& Total & 0.330 & 0.334 & 0.333 & 99.000 & 100.240 \\
	\midrule
	& 0 & 0.361 & 0.361 & 0.361 & 99.995 & 100.001  \\
	& 1 & 0.069 & 0.069 & 0.069 & 99.664 & 99.986  \\
	\textit{Betula pendula} ($\mathrm{ST}_2$) & 2 & 0.002 & 0.002 & 0.002 & 74.402 & 100.380  \\
	& 3 & -- & -- & -- & --      & -- \\
	& Total & 0.431 & 0.431 & 0.431 & 99.826 & 100.000 \\
	\midrule
	& 0 & 0.290 & 0.290 & 0.291 & 99.874 & 99.905  \\
	& 1 & 0.160 & 0.168 & 0.168 & 95.263 & 100.018  \\
	\textit{Populus nigra} ($\mathrm{ST}_3$) & 2 & 0.029 & 0.036 & 0.036 & 81.669 & 100.388  \\
	& 3 & 0.004 & 0.005 & 0.004 & 84.422 & 102.212 \\
	& Total & 0.483 & 0.499 & 0.499 & 96.880 & 99.998 \\
	\midrule
	& 0 & 0.054 & 0.054 & 0.054 & 100.000 & 100.000  \\
	& 1 & 0.008 & 0.008 & 0.008 & 94.874 & 99.943  \\
	\textit{Populus tremuloides} ($\mathrm{ST}_4$) & 2 & -- & -- & -- & -- & --  \\
	& 3 & -- & -- & -- & -- & -- \\
	& Total & 0.062 & 0.063 & 0.063 & 99.312 & 99.992 \\
	\midrule
	& 0 & 0.487 & 0.487 & 0.487 & 99.951 & 99.953  \\
	& 1 & 0.080 & 0.086 & 0.089 & 88.917 & 96.613  \\
	\textit{Ailanthus fordii} ($\mathrm{RT}_1$)  & 2 & 0.036 & 0.049 & 0.048 & 73.679 & 100.231  \\
	& 3 & 0.005 & 0.008 & 0.008 & 62.377 & 100.461 \\
	& Total & 0.607 & 0.630 & 0.633 & 95.894 & 99.509 \\
	\midrule
	& 0 & 0.389 & 0.389 & 0.389 & 100.000 & 100.000  \\
	& 1 & 0.411 & 0.420 & 0.431 & 95.496 & 97.501  \\
	\textit{Delonix regia} ($\mathrm{RT}_2^{(1)}$) & 2 & 0.170 & 0.218 & 0.214 & 79.547 & 101.910  \\
	& 3 & 0.143 & 0.177 & 0.171 & 83.764 & 103.680 \\
	& Total & 1.113 & 1.204 & 1.204 & 92.456 & 99.967\\
	\midrule
	& 0 & 0.362 & 0.362 & 0.362 & 100.000 & 100.000  \\
	& 1 & 0.254 & 0.254 & 0.254 & 100.000 & 100.000  \\
	\textit{Delonix regia} ($\mathrm{RT}_2^{(2)}$) & 2 & 0.255 & 0.282 & 0.282 & 90.433 & 99.926  \\
	& 3 & 0.265 & 0.338 & 0.338 & 78.383 & 100.097 \\
	& Total & 1.136 & 1.236 & 1.236 & 91.906 & 100.010 \\
	\midrule
	& 0 & 0.044 & 0.044 & 0.044 & 100.000 & 100.000  \\
	& 1 & 0.013 & 0.013 & 0.013 & 99.859 & 99.981  \\
	\textit{Crateva unilocularis} ($\mathrm{RT}_3^{(1)}$) & 2 & 0.016 & 0.016 & 0.016 & 99.022 & 100.019  \\
	& 3 & 0.030 & 0.032 & 0.032 & 95.614 & 100.002 \\
	& Total & 0.103 & 0.105 & 0.105 & 98.511 & 100.001 \\
	\midrule
	& 0 & 0.038 & 0.038 & 0.038 & 100.000 & 100.000  \\
	& 1 & 0.010 & 0.010 & 0.010 & 99.944 & 99.949  \\
	\textit{Crateva unilocularis} ($\mathrm{RT}_3^{(2)}$) & 2 & 0.008 & 0.008 & 0.008 & 98.305 & 100.059  \\
	& 3 & 0.015 & 0.018 & 0.018 & 87.807 & 100.004 \\
	& Total & 0.071 & 0.073 & 0.073 & 96.887 & 100.001 \\
	\midrule 
	& 0 & 0.038 & 0.038 & 0.038 & 100.000 & 100.000  \\
	& 1 & 0.004 & 0.004 & 0.004 & 99.601 & 99.694  \\
	\textit{Crateva unilocularis} ($\mathrm{RT}_3^{(3)}$) & 2 & 0.006 & 0.006 & 0.006 & 97.695 & 99.730  \\
	& 3 & 0.013 & 0.016 & 0.016 & 81.330 & 100.088 \\
	& Total & 0.060 & 0.063 & 0.063 & 95.167 & 99.978 \\
	\bottomrule
\end{tabular}
%	\vspace{0.5em}
%\begin{minipage}{6.7in}
%	\small *Volumes and totals were computed from unrounded values; displayed level-wise volumes may not sum exactly to the displayed total because of rounding.
%\end{minipage}
\end{table*}

\subsubsection{Quantitative performance evaluation}

Table~\ref{tab:Performance metrics} summarizes the five evaluation metrics for each test tree. The proposed method shows close agreement with the selected reference QSMs under the evaluated protocol. The reported volume errors should be interpreted as reference-QSM agreement rather than absolute physical-volume accuracy.

The m$\mathrm{E_V}$ across all test trees is 0.081\%, indicating that the completed QSM preserves total woody volume with an average relative error below one-thousandth. The best performance is observed for \textit{Betula pendula} ($\mathrm{ST}_2$) and \textit{Crateva unilocularis} ($\mathrm{RT}_3^{(2)}$), both with $\mathrm{E_V} < 0.001$\%, where the partial input is already highly complete. For \textit{Ailanthus fordii} ($\mathrm{RT}_1$), whose partial QSM contained 95.894\% of the reference-QSM volume, the completed QSM had a relative reference-volume error of $\mathrm{E_V} = 0.491\%$. This result demonstrates close agreement under the present within-dataset evaluation, but does not by itself establish generalization to unseen species, sites, or acquisition systems.

The m$\mathrm{E_{LV}}$ is 0.676\% on the evaluated test set, indicating close agreement with the reference volumes aggregated by branch level. This validates the design of the level-wise volume loss, which explicitly supervises per-level volume distribution.

The m$\mathrm{B_V}$ is $-0.030$\%, confirming that the method is nearly unbiased: it neither systematically over-generates nor under-generates volume. Across individual trees, $\mathrm{B_V}$ ranges from $-0.491$\% for \textit{Ailanthus fordii} ($\mathrm{RT}_1$) to $+0.240$\% for \textit{Acer palmatum} ($\mathrm{ST}_1$). The narrow range indicates that the volume budget mechanism effectively scales the generated volume to match the complete reference.

The \(\mathrm{MVR}_{\mathrm{agg}}\) is 98.715\%, indicating that the net added volume closely matches the aggregate reference missing volume. For \textit{Acer palmatum} ($\mathrm{ST}_1$), the MVR of 124.025\% indicates that the generated volume exceeded the reference missing volume by approximately 24\%. This ratio is amplified by the small missing-volume denominator and should not be interpreted as improved completion performance. For trees with substantial missing volume (e.g., \textit{Delonix regia} ($\mathrm{RT}_2^{(1)}$) and \textit{Delonix regia} ($\mathrm{RT}_2^{(2)}$) with Par/Com ratio of 92.456\% and 91.906\%), the model recovers 99.563\% and 100.119\% of the missing volume, respectively.

The $\mathrm{CD_{ske}}$ ranges from 0.018 (\textit{Crateva unilocularis} ($\mathrm{RT}_3^{(1)}$)) to 0.245 (\textit{Delonix regia} ($\mathrm{RT}_2^{(1)}$)), with a mean value of 0.113. The largest errors occur in structurally complex trees with extensive branching, where the partial input contains limited topological information. Nevertheless, the relatively low \(\mathrm{CD_{ske}}\) values indicate that the completed skeleton maintains geometric fidelity to the complete reference.

The method performs consistently across both simulated and real tree species. The simulated species (\textit{Acer palmatum} ($\mathrm{ST}_1$), \textit{Betula pendula} ($\mathrm{ST}_2$), \textit{Populus nigra} ($\mathrm{ST}_3$), \textit{Populus tremuloides} ($\mathrm{ST}_4$)) show slightly better metrics compared to real species, likely due to the absence of leaf noise, more controlled acquisition conditions in simulation, simple and regular crown structure, and lower occlusion.

Overall, the quantitative results indicate low volume errors and a small mean volume bias for the evaluated trees and acquisition settings.

\begin{table*}[H]
	\centering
	\caption{Completion performance on the simulated and real TLS test sets. Lower is better for errors; closer to 0 is better for volume bias.}
	\label{tab:Performance metrics}
	\begin{tabular}{lccccc}
		\toprule
		Tree name & \(\mathrm{E_V}\) (\%) $\downarrow$ & \(\mathrm{E_{LV}}\) (\%) $\downarrow$ & \(\mathrm{B_V}\) (\%)  & $\mathrm{MVR}$ (\%)  & \(\mathrm{CD_{ske}}\) (m) $\downarrow$ \\
		\midrule
		\textit{Acer palmatum} ($\mathrm{ST}_1$)  & 0.240 & 2.671 & $+0.240$ & 124.025 & 0.088 \\
		\textit{Betula pendula} ($\mathrm{ST}_2$) & 0.000 & 0.099 & $+0.000$ & 100.168 & 0.135 \\
		\textit{Populus nigra} ($\mathrm{ST}_3$)  & 0.002 & 0.679 & $-0.002$ & 99.928 & 0.144 \\
		\textit{Populus tremuloides} ($\mathrm{ST}_4$) & 0.008 & 0.014 & $-0.008$ & 98.894 & 0.097 \\
		\textit{Ailanthus fordii} ($\mathrm{RT}_1$)     & 0.491 & 1.031 & $-0.491$ & 88.046 & 0.121 \\
		\textit{Delonix regia} ($\mathrm{RT}_2^{(1)}$)	   & 0.033 & 2.022 & $-0.033$ & 99.563 & 0.245 \\
		\textit{Delonix regia} ($\mathrm{RT}_2^{(2)}$)	   & 0.010 & 0.043 & $+0.010$ & 100.119 & 0.196 \\
		\textit{Crateva unilocularis} ($\mathrm{RT}_3^{(1)}$) & 0.001 & 0.010 & $+0.001$ & 100.075 & 0.018 \\
		\textit{Crateva unilocularis} ($\mathrm{RT}_3^{(2)}$) & 0.001 & 0.029 & $+0.001$ & 100.029 & 0.020 \\
		\textit{Crateva unilocularis} ($\mathrm{RT}_3^{(3)}$) & 0.022 & 0.166 & $-0.022$ & 99.542 & 0.061 \\
%		\midrule
%		Mean of above trees					     & 0.081 & 0.676 & $-0.030$ & 98.715 & 0.113 \\
		\bottomrule
	\end{tabular}
\end{table*}

\subsection{Comparison with baselines}
Table~\ref{tab:metrics_tradition} shows that the partial QSM baseline produces m$E_{V}$ of 3.416\% and m$\mathrm{E_{LV}}$ of 6.767\%, with a negative volume bias of $-3.416$\%, confirming that missing branches lead to systematic volume underestimation. The regression-based baseline reduces the volume errors to 2.104\% and 4.579\%, respectively, and achieves the $\mathrm{MVR_{agg}}$ of 95.090\%. This indicates that allometric correction can recover part of the missing volume, but it does not reconstruct the missing branch geometry and therefore cannot provide a skeleton-distance result. In comparison, the proposed method achieved substantially lower volume errors and near-zero bias while explicitly completing the tree structure. ArborNet reduces volume error relative to the partial QSM but slightly increases mean skeleton Chamfer distance from 0.104 m to 0.113 m. Thus, improved volumetric completeness is not accompanied by an improvement in the evaluated global centreline metric.

Table~\ref{tab:metrics_tradition} also compares the proposed method against three state-of-the-art point cloud completion methods (PoinTr, AdaPoinTr, and SnowflakeNet) followed by TreeQSM reconstruction. The evaluated point-completion–to–TreeQSM pipelines produce larger volume errors than ArborNet. Their mean signed volume biases are positive, ranging from +35.305\% to +57.152\%, but individual trees can exhibit substantial underestimation. These results reflect the combined effects of point generation and subsequent QSM reconstruction rather than point-completion quality alone. In contrast, the proposed QSM-domain method achieves m$\mathrm{B_V} = -0.030$\%, representing a bias reduction of over 35 percentage points compared to the best point-based baseline. The m$\mathrm{E_V}$ of the proposed method is 0.081\%, while the baselines have errors exceeding 60\%. The proposed method achieves \(\mathrm{MVR_{agg}}\) = 98.715\%, accurately matching the missing volume of the reference QSM without significant under- or over-generation. Point completion baselines show m$\mathrm{E_{LV}}$ values of approximately 200\%, indicating that the generated volume is poorly distributed across branch levels. This is expected, as point completion methods have no mechanism to enforce hierarchical branch-level constraints. The proposed method achieves m$\mathrm{E_{LV}} = 0.676$\%, demonstrating that volume is correctly allocated across levels. The m\(\mathrm{CD_{ske}}\) for point completion baselines ranges from 0.341 to 0.369, approximately three times larger than the proposed method (0.113). This indicates that the skeletal geometry generated by point completion methods deviates substantially from the reference QSM.

\begin{figure*}[!ht]
	\centering
	\includegraphics[width=0.9\linewidth]{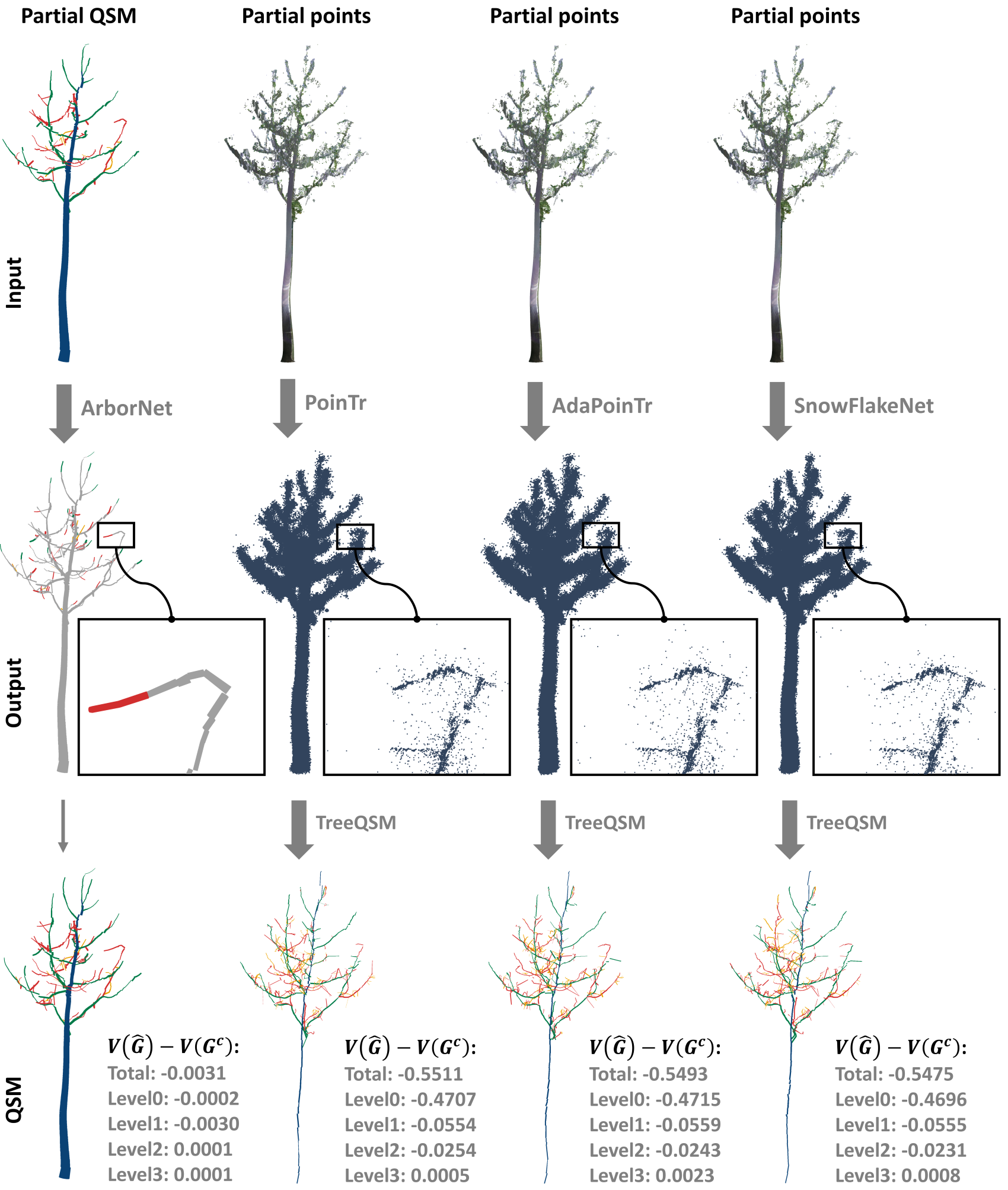}
	\caption{Qualitative results comparison of point cloud completion-based methods and proposed method. Selected details are marked with bold outlines.}
	\label{fig:Results_tradition}
\end{figure*}

Figure~\ref{fig:Results_tradition} illustrates the visual differences between methods. Point completion methods produce smooth, dense point clouds that appear visually plausible, but the subsequent TreeQSM reconstruction converts them into unnatural cylinder arrangements with incorrect topology, overly thick or thin branches, and missing fine structures. The substantial stem-volume underestimation in the illustrated example does not necessarily indicate removal of the observed stem points. TreeQSM refits the observed and generated points together rather than retaining the original stem cylinders. Generated points inside or outside the woody surface can affect point grouping and cylinder fitting, potentially producing underestimated stem radii. Because volume depends strongly on radius, this can result in a large volume deficit. This is a plausible explanation of the observed reconstruction behaviour, rather than an isolated causal finding. In contrast, the proposed method generates structurally coherent QSMs that preserve branch topology, radius tapering, and hierarchical organization. QSM-domain completion, with its explicit geometry, skeleton, and volume constraints, provides a more appropriate solution.

\begin{table*}[H]
	\centering
	\caption{The mean metric values across all test trees for different methods. Lower is better for errors; closer to 0 is better for volume bias.}
	\label{tab:metrics_tradition}
	\begin{tabular}{lccccc}
		\toprule
		Method & m\(\mathrm{E_V}\) (\%) $\downarrow$ & m\(\mathrm{E_{LV}}\) (\%) $\downarrow$ & m\(\mathrm{B_V}\) (\%)  & $\mathrm{MVR_{agg}}$ (\%)  & m\(\mathrm{CD_{ske}}\) (m) $\downarrow$ \\
		\midrule
		Partial QSM (TreeQSM) & 3.416 & 6.767 & $-3.416$ & 0.0 & 0.104 \\
		Regression-based baseline & 2.104 & 4.579 & $-2.104$ & 95.090 & N/A \\
		PoinTr \citep{yu2021pointr} + TreeQSM & 84.667 & 206.511 & $+55.799$ & 891.583 & 0.341 \\
		AdaPoinTr \citep{10232862} + TreeQSM & 80.603 & 195.920 & $+57.152$ & 1002.482 & 0.352 \\
		SnowflakeNet \citep{xiang2021snowflakenet} + TreeQSM & 62.853 & 188.948 & $+35.305$ & 631.253 & 0.369 \\
		Proposed ArborNet & 0.081 & 0.676 & $-0.030$ & 98.715 & 0.113 \\
		\bottomrule
	\end{tabular}
\end{table*}

\subsection{Ablation study}
To assess the contribution of each loss component, we compare the full model against six ablated variants. These include removing the total volume loss, removing the level-wise volume loss, removing both volume losses simultaneously, removing the branch-start loss, removing the axis smoothness prior, and removing the radius prior. Each variant is trained under identical settings to the full model, and the resulting metrics are reported in Table~\ref{tab:ablation}, where ``w/o level volume” removes $\Lc_\mathrm{lvol}$ and its jointly weighted global-head supervision $\Lc_\mathrm{gvol}$ and ``w/o total + level volume” additionally removes 
$\Lc_\mathrm{vol}$. The anchor budget loss remains active in both variants. The results reveal the contribution of each loss term to overall performance.

\begin{figure*}[!htH]
	\centering
	\includegraphics[width=0.9\linewidth]{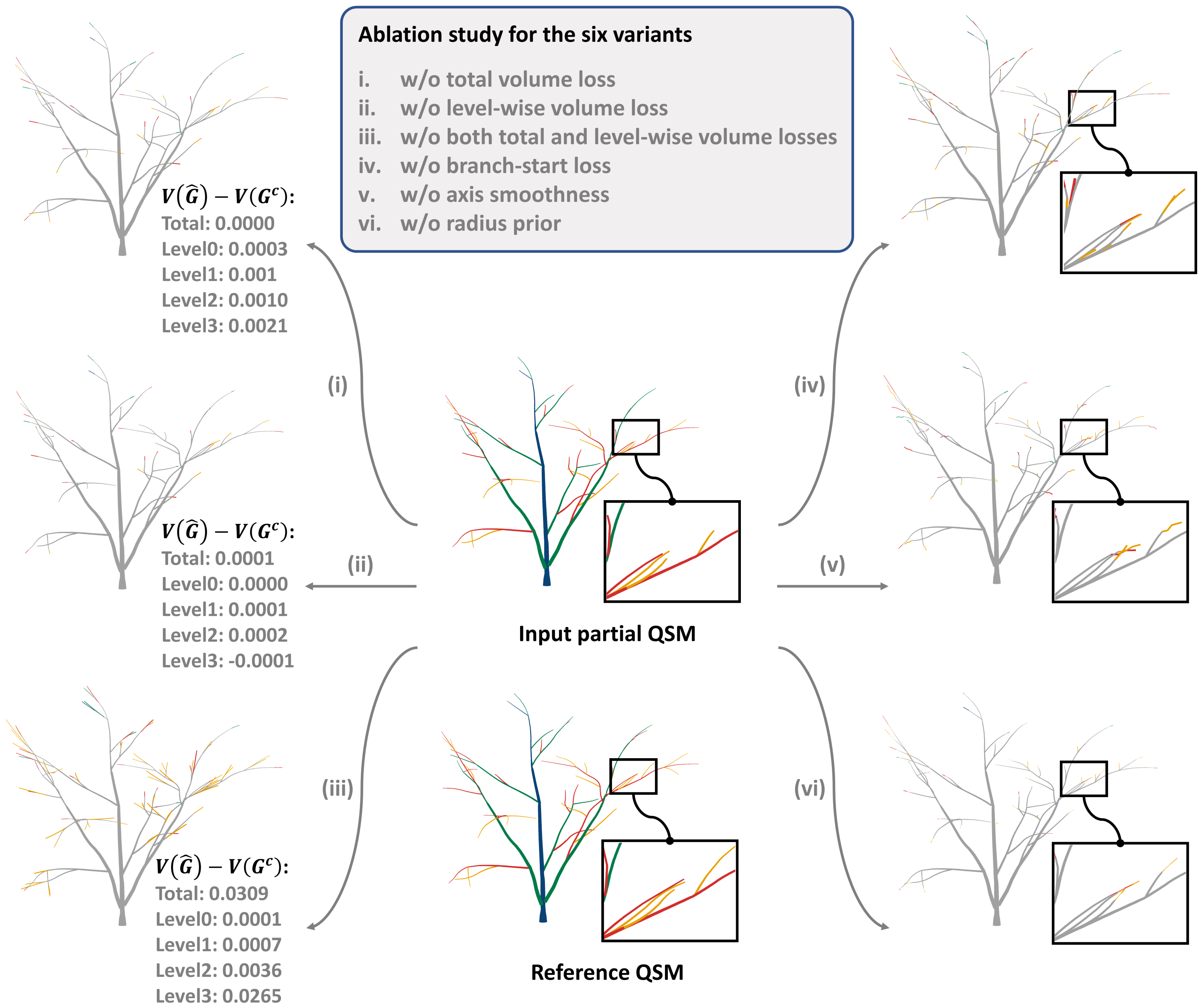}
	\caption{Ablation study for the six variants.}
	\label{fig:Results_ablation}
\end{figure*}

Variants without volume losses produce inconsistent completions with either under-generated or over-generated volume. The variant without the radius prior shows noticeably thicker or thinner branches compared to the reference QSM. Figure~\ref{fig:Results_ablation} provides visual evidence complementing the results. The full model shown in Figure~\ref{fig:Results_sim} ($\mathrm{ST}_1$) produces the most visually coherent result, with generated cylinders smoothly integrating into the existing structure.

Removing the total volume loss alone ($\lambda_\mathrm{vol}=0$) increases m$\mathrm{E_V}$ from 0.081\% to 2.353\%, indicating that the total volume loss effectively constrains the global magnitude of generated volume. Removing the level-wise volume loss alone ($\lambda_\mathrm{lvol}=0$) increases m$\mathrm{E_V}$ to 2.826\% and degrades m$\mathrm{E_{LV}}$ from 0.676\% to 8.766\%, confirming that per-level supervision is essential for distributing volume correctly across branch orders. Joint removal of the total- and level-wise volume terms increases m$\mathrm{E_V}$ to 64.777\% and m$\mathrm{B_V}$ to +64.777\%, indicating substantial overestimation under this ablation setting. The m\(\mathrm{CD_{ske}}\) also increases to 0.246, more than double the full model. Following Equation (\ref{eq:28}), setting $\lambda_\mathrm{lvol}=0$ removes both the generated level-wise volume loss and the global budget-head loss. The joint-removal variant sets $\lambda_\mathrm{vol}=\lambda_\mathrm{lvol}=0$, while retaining the anchor budget loss with $\lambda_\mathrm{bud}=0.1$. Consequently, the global budget head receives no direct budget supervision in this variant, whereas the local budget head remains supervised.

Removing branch-start supervision ($\lambda_\mathrm{start}=0$) reduces $\mathrm{MVR_{agg}}$ from 98.715\% to 40.188\% and increases m$\mathrm{E_V}$ to 2.671\%. The negative mean volume bias indicates net underestimation, although individual trees may exhibit either under- or over-generation.

Removing axis smoothness ($\lambda_\mathrm{smooth}=0$) increases m$\mathrm{E_V}$ to 2.431\% and degrades m$\mathrm{E_{LV}}$ to 4.245\%. The m\(\mathrm{CD_{ske}}\) increases from 0.113 to 0.205, suggesting that without smoothness regularization, generated branches become geometrically irregular, affecting both volume and topological quality.

Removing the radius prior ($\lambda_\mathrm{rad}=0$) yields significant degradation: m$\mathrm{E_V}$ increases to 3.204\%, m$\mathrm{E_{LV}}$ increases to 5.662\%, and m$\mathrm{B_V}$ drops to $-2.602$\%. This demonstrates that the radius tapering constraint is critical for maintaining biologically plausible radii and accurate volume estimation. Without this prior, the model tends to generate cylinders with unrealistic radius profiles.

The full model provides the lowest total- and level-wise volume errors, but it does not minimize skeleton Chamfer distance among all ablated variants. Removing total-volume or branch-start supervision slightly lowers this geometric metric while substantially worsening volume recovery, indicating a trade-off between volumetric agreement and the evaluated centreline distance.

\begin{table*}[H]
\centering
\caption{Mean values of metrics for all test datasets based on ablation study. Lower is better for errors; closer to 0 is better for volume bias.}
\label{tab:ablation}
\begin{tabular}{lccccc}
	\toprule
	Variant & m\(\mathrm{E_V}\) (\%) $\downarrow$ & m\(\mathrm{E_{LV}}\) (\%) $\downarrow$ & m\(\mathrm{B_V}\) (\%)  & $\mathrm{MVR_{agg}}$ (\%)  & m\(\mathrm{CD_{ske}}\) (m) $\downarrow$ \\
	\midrule
	Full model & 0.081 & 0.676 & $-0.030$ & 98.715 & 0.113  \\
	w/o total volume & 2.353 & 3.790 & $-0.138$ & 55.512 & 0.106\\
	w/o level volume & 2.826 & 8.766 & $+1.819$ & 106.482 & 0.120\\
	w/o total + level volume & 64.777 & 67.584 & $+64.777$ & 378.138 & 0.246\\
	w/o branch-start & 2.671 & 4.706 & $-0.449$ & 40.188 & 0.104\\
	w/o axis smoothness & 2.431 & 4.245 & $-0.589$ & 40.792 & 0.205\\
	w/o radius prior & 3.204 & 5.662 & $-2.602$ & 11.727 & 0.113\\
	\bottomrule
\end{tabular}
\end{table*}

\subsection{Sensitivity analysis}
\begin{figure*}[!ht]
	\centering
	\includegraphics[width=0.9\linewidth]{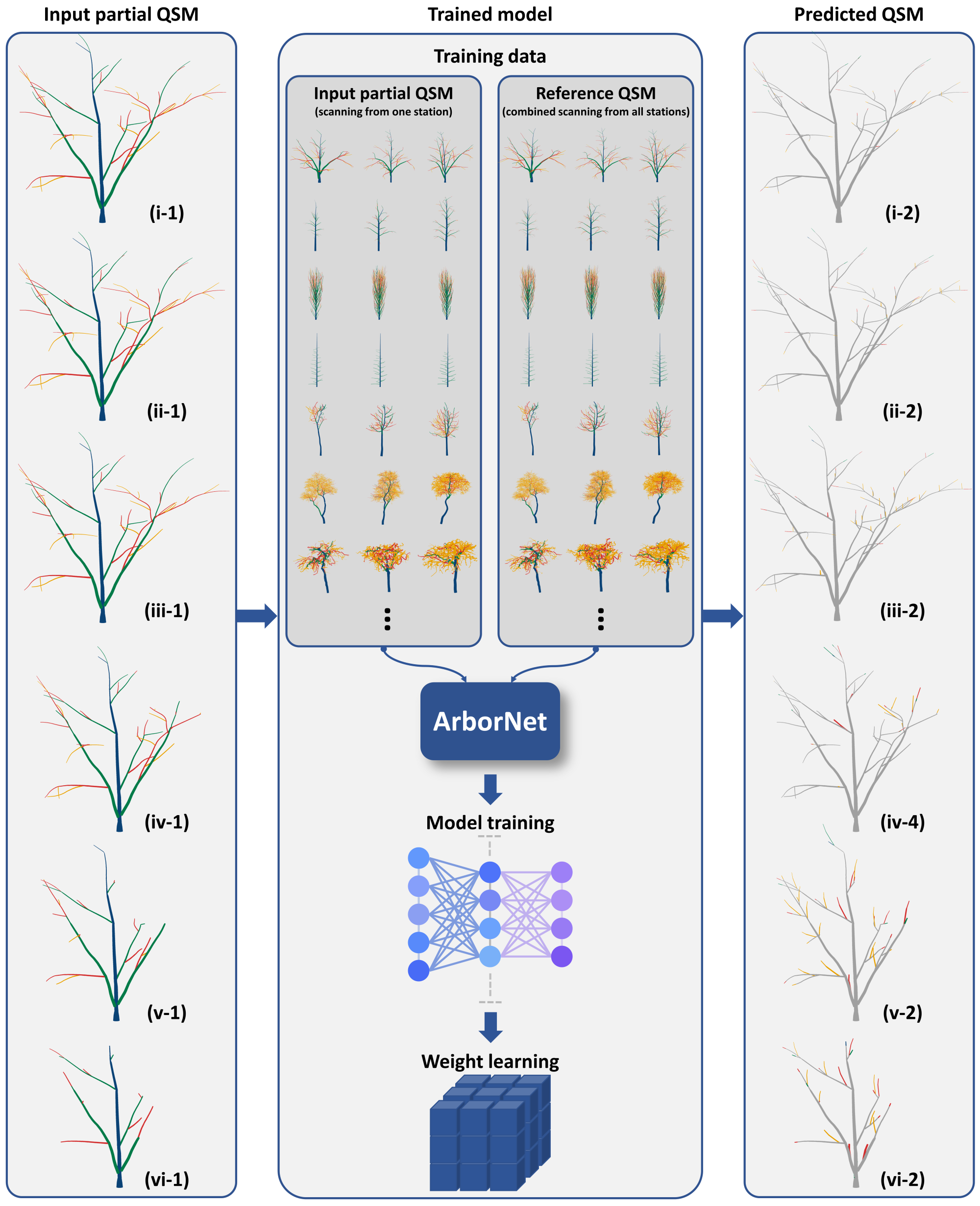}
	\caption{Completion results of sensitivity analysis with input partial QSMs and predicted QSMs. (i-1) and (i-2): 3 setups with scanning resolution of $0.02^\circ$, (ii-1) and (ii-2): 2 setups with scanning resolution of $0.02^\circ$, (iii-1) and (iii-2): 1 setup with scanning resolution of $0.02^\circ$, (iv-1) and (iv-2): scanning resolution of $0.04^\circ$ with 1 setup, (v-1) and (v-2): scanning resolution of $0.06^\circ$ with 1 setup, (vi-1) and (vi-2): and scanning resolution of $0.08^\circ$ with 1 setup.}
	\label{fig:Results_sensitivity}
\end{figure*}

Table~\ref{tab:sentivility} and Figure~\ref{fig:Results_sensitivity} present the sensitivity analysis of the proposed method to the number of TLS scan positions and the scanning resolution (zenith and azimuth) in the DART parameter setting.

As the number of TLS setups decreases with fixed resolution zenith and azimuth of $0.02^\circ$ (see Table \ref{tab:DART_scene_configuration}), the partial input becomes increasingly incomplete, yet the proposed method maintains robust performance. The reported m$\mathrm{E_V}$ is 0.108\%, 0.082\%, and 0.063\% for three, two, and one scan positions, respectively. Thus, total-volume error does not increase monotonically as the number of scan positions decreases, while, m$\mathrm{E_{LV}}$ increases from 0.362\%, 0.431\% to 0.866\% for three, two, and one scan positions. The m\(\mathrm{CD_{ske}}\) increases from 0.034 (3 setups) to 0.116 (1 setup), reflecting the increased geometric uncertainty when fewer views are available.

Notably, the performance degradation from 3 to 1 setup is relatively small compared to the loss of input information. This suggests that the proposed method effectively leverages learned structural priors to compensate for missing observations, even when two TLS views are removed from the three.

Degrading scanning resolution has a more pronounced effect than reducing the number of setups. Keeping a single setup of TLS with both zenith and azimuth resolution decreasing from $0.02^\circ$ to $0.04^\circ$, m$\mathrm{E_V}$ increases from 0.063\% to 0.792\% and m\(\mathrm{CD_{ske}}\) increases from 0.116 to 0.292. At $0.06^\circ$, m$\mathrm{E_V}$ reaches 3.775\% and $\mathrm{MVR_{agg}}$ drops to 76.169\%. At $0.08^\circ$, performance deteriorates substantially: m$\mathrm{E_V} = 11.499\%$, m$\mathrm{B_V} = -11.499\%$, and m\(\mathrm{CD_{ske}}\) exceeds 1.0.

This sensitivity to scanning resolution is expected, because lower-resolution scans reduce point density on fine branches, making it difficult to accurately capture structural topology and radius. The method relies on accurate TreeQSM inputs, which in turn depend on sufficient point density. For practical applications, these results suggest that the proposed method is effective with standard TLS resolution settings ($\leq 0.04^\circ$) and can tolerate moderate resolution degradation. However, extremely coarse scanning ($\geq 0.08^\circ$) may require additional denoising or preprocessing.

The sensitivity analysis provides practical guidance for TLS data collection. The number of TLS positions can be reduced from three to one with only a small loss in completion quality, making single-station acquisition a viable option when combined with the proposed completion method. However, scanning resolution should be maintained at $\leq 0.04^\circ$ to ensure sufficient detail for QSM reconstruction and subsequent completion.

\begin{table*}[H]
	\centering
	\caption{Mean values of metrics for all the simulated test datasets based on sensitivity analysis. Lower is better for errors; closer to 0 is better for volume bias.}
	\label{tab:sentivility}
	\begin{tabular}{lccccc}
		\toprule
		Variant & m\(\mathrm{E_V}\) (\%) $\downarrow$ & m\(\mathrm{E_{LV}}\) (\%) $\downarrow$ & m\(\mathrm{B_V}\) (\%)  & $\mathrm{MVR_{agg}}$ (\%)  & m\(\mathrm{CD_{ske}}\) (m) $\downarrow$ \\
		\midrule
		3 setups (res. $0.02^\circ$) & 0.108 & 0.362 & $-0.108$  & 74.240 & 0.034  \\
		2 setups (res. $0.02^\circ$) & 0.082 & 0.431 & $+0.079$ & 119.143 & 0.055\\
		1 setup (res. $0.02^\circ$)  & 0.063 & 0.866 & $+0.058$ & 103.914 & 0.116 \\
		res. $0.04^\circ$ (1 setup)  & 0.792 & 8.184 & $-0.792$ & 87.668 & 0.292 \\
		res. $0.06^\circ$ (1 setup)  & 3.775 & 16.666 & $-3.775$ & 76.169 & 0.630\\
		res. $0.08^\circ$ (1 setup)  & 11.499 & 35.578 & $-11.499$ & 42.246 & 1.001\\
		\bottomrule
	\end{tabular}
\end{table*}

\subsection{Perturbation test}
To assess sensitivity to input partial QSM uncertainty, we apply controlled perturbations to the geometric parameters of the input partial QSMs. Cylinder radii and lengths are perturbed by zero-mean Gaussian noise with standard deviations of 5\% and axis directions of cylinders are perturbed by 5 degrees. Each setting is repeated 10 times, and Table~\ref{tab:perturbation} summarizes the resulting changes in volume agreement and skeleton geometry. It summarizes the results of ArborNet to perturbations in the input partial QSM cylinder parameters. Radius perturbation primarily affects volumetric performance, increasing m\(\mathrm{E_V}\) from 0.081\% to 0.089\% and m\(\mathrm{E_{LV}}\) from 0.676\% to 0.684\%, while producing little change in skeleton geometry. In contrast, perturbing cylinder axis directions substantially increased m\(\mathrm{CD_{ske}}\) from 0.113 m to 0.239 m, although its effect on the volume-based metrics remained limited. Cylinder-length perturbation had a negligible effect on all evaluated metrics, with m\(\mathrm{E_V}\), m\(\mathrm{E_{LV}}\), and m\(\mathrm{CD_{ske}}\) remaining nearly unchanged. $\mathrm{MVR_{agg}}$ increases from 98.715\% without perturbation to 104.051\%, 101.030\%, and 102.047\% under radius, length, and axis perturbations, respectively. These increases should not be interpreted as improved recovery as MVR is sensitive to changes in the partial-volume denominator, and the volume-error metrics remain the more direct measures of agreement with the fixed reference QSMs. Repeated TreeQSM reconstructions show that run-to-run variability is mainly reflected in cylinder subdivision and length, whereas radius and axis direction are comparatively stable. Therefore, under the tested perturbation settings, the parameter variability typically introduced by repeated TreeQSM reconstruction has only a limited effect on ArborNet’s completion performance. Nevertheless, larger errors in radius or axis direction could affect volume prediction and skeleton geometry, respectively.

\begin{table*}[H]
	\centering
	\caption{Mean values of metrics for all test datasets for perturbation test. Lower is better for errors; closer to 0 is better for volume bias.}
	\label{tab:perturbation}
	\begin{tabular}{lccccc}
		\toprule
		Perturbation item & m\(\mathrm{E_V}\) (\%) $\downarrow$ & m\(\mathrm{E_{LV}}\) (\%) $\downarrow$ & m\(\mathrm{B_V}\) (\%)  & $\mathrm{MVR_{agg}}$ (\%)  & m\(\mathrm{CD_{ske}}\) (m) $\downarrow$ \\
		\midrule
		N/A				& 0.081 & 0.676 & $-0.030$ & 98.715 & 0.113  \\
		Cylinder radii  & 0.089 & 0.684 & $-0.039$ & 104.051 & 0.114\\
		Cylinder lengths & 0.081 & 0.677 & $-0.031$ & 101.030 & 0.113\\
		Cylinder axis directions & 0.084 & 0.681 & $-0.043$ & 102.047 & 0.239\\
		\bottomrule
	\end{tabular}
\end{table*}

%\subsection{Effect of constrained decoding}
%
%\todo{If available, add an experiment comparing unconstrained decoding, probability-only decoding, and volume-budgeted constrained decoding.}
%
%\begin{table*}[htbp]
%\centering
%\caption{Effect of constrained decoding.}
%\label{tab:decoding}
%\begin{tabular}{lrrrr}
%\toprule
%Decoding strategy & \(\mathrm{E_V}\) (\%) & \(\mathrm{E_{LV}}\) (\%) & Over-generation rate & Visual plausibility \\
%\midrule
%Probability only & \todo{} & \todo{} & \todo{} & \todo{} \\
%+ axis/radius constraints & \todo{} & \todo{} & \todo{} & \todo{} \\
%+ level-wise volume budget & \todo{} & \todo{} & \todo{} & \todo{} \\
%Full constrained decoding & \todo{} & \todo{} & \todo{} & \todo{} \\
%\bottomrule
%\end{tabular}
%\end{table*}

% ============================================================
\section{Discussion} \label{sec:6}
% ============================================================

\subsection{Interpreting the comparison with point-completion methods}
The proposed method differs fundamentally from point cloud completion. Point-based methods infer a set of 3D points, and structural attributes must be recovered through a separate QSM reconstruction step. Errors in generated point distributions can propagate into QSM fitting, causing incorrect branch topology, unrealistic radii, and biased volume. Several characteristics of the evaluated point-based solutions may contribute to their downstream QSM errors, including the absence of explicit cylinder-volume objectives and the sensitivity of QSM fitting to generated point distributions.

First, point completion methods optimize geometric distances between point sets (e.g., Chamfer distance) but do not explicitly enforce structural constraints. A completed point cloud may achieve low Chamfer distance while containing topologically incorrect or biologically implausible structures. When such point clouds are passed through TreeQSM, the reconstruction algorithm fits cylinders to the dense, unconstrained point distributions, often producing overly thick or thin, spurious branches that inflate or deflate volume estimates.

Second, point completion methods lack hierarchical awareness. They do not distinguish between trunk, primary branches, secondary branches, and twigs. The evaluated point-based methods do not explicitly optimize branch-order QSM volumes, so agreement in point geometry does not necessarily ensure agreement in the reconstructed branch-order volumes.

Third, point completion methods do not incorporate radius tapering or growth constraints. Without these priors, generated points can produce arbitrary radius profiles during QSM reconstruction, violating the pipe model theory and resulting in unrealistic volume estimates.

In contrast, the proposed framework predicts QSM cylinders directly. Each generated element has explicit radius, length, axis, parent relation, and branch level. Therefore, the completion target aligns with the structural quantities needed in downstream ecological applications and avoids the error propagation inherent in point-to-cylinder conversion.

\subsection{Importance of volume-aware learning}
Since woody volume depends quadratically on radius and linearly on length, a small error in radius can cause a disproportionately large error in volume. This implies that visually small geometric deviations in large-radius branches may produce substantial volumetric errors, while visually noticeable errors in fine twigs may have limited volumetric impact. Consequently, completion methods that optimize only geometric proximity, such as Chamfer distance, may achieve visually plausible results while producing severely biased volume estimates. 

To address this, the proposed framework integrates volume information into both training and inference. During training, total-volume and level-wise volume losses constrain the global magnitude and hierarchical distribution of generated volume, respectively, whereas the global volume head loss and the anchor budget loss train the two budget predictors that provide inference-time volume constraints. During inference, the predicted level-wise missing volume budget acts as a hard constraint: candidate cylinders are accepted only if the accumulated volume at their level remains within the budget. The checks limit accepted volume relative to the predicted budgets. They do not guarantee agreement with reference volume, nor do upper-bound constraints alone prevent under-generation. Ablation results confirm that removing volume-aware losses severely degrades volumetric accuracy, demonstrating the necessity of this design.

Furthermore, it is worth noting that the global volume budget does not compromise the geometric fidelity of the generated tree skeleton. As evidenced by the ablation study in Table 8, when the volume losses are fully integrated (Full model), the skeleton Chamfer distance (m\(\mathrm{CD_{ske}}\)) remains at a highly precise level of \(0.113\text{ m}\). In contrast, removing both total and level-wise volume losses causes the geometric error to more than double (\(0.246\text{ m}\)). The joint removal of the global and hierarchical volume losses increases both the volume bias and the skeleton distance. However, neither individual removal produces a consistent increase in the skeleton distance. These results support the role of the volume losses in controlling the generated volume, but do not isolate their effect on the spatial placement of newly generated cylinders.

\subsection{Importance of skeleton-aware learning}
Tree woody structures are not merely collections of spatial points, but connected branching systems whose geometry depends on parent–child relationships and branch continuity. Geometric proximity alone does not explicitly represent these relationships. Consequently, a completion that produces a plausible point distribution may still yield disconnected elements or inconsistent branch organization when converted into a QSM. Incorporating the observed skeleton into learning provides a direct representation of the structural relationships relevant to cylinder-based completion.

The proposed framework uses skeleton information during both feature extraction and decoding. The multi-relation graph encoder aggregates information through parent, child, and intra-branch edges, allowing each observed cylinder to be represented in the context of its neighbouring woody structure. The anchor-conditioned decoder then separates missing-structure generation into terminal continuation and branch-start generation. Generated cylinders are attached to their parent endpoints, while branch levels are assigned according to the corresponding parent–child rules. These mechanisms enforce attachment and branch-order consistency by construction, rather than relying on a point-to-cylinder reconstruction step. The axis-smoothness and radius priors further regularize the geometry of generated elements.

The ablation results provide partial support for these structural constraints. Removing axis smoothness increases m\(\mathrm{CD_{ske}}\) from 0.113 m to 0.205 m, suggesting that directional regularization contributes to centreline agreement with the reference QSMs. Removing branch-start supervision reduces $\mathrm{MVR_{agg}}$ from 98.715\% to 40.188\%, indicating its importance for recovering missing structures under the evaluated setting. However, these experiments do not isolate the contribution of the graph encoder or establish the accuracy of recovered branching topology. Skeleton-aware learning should therefore be understood as the explicit use of observed structural relationships to guide completion, rather than a guarantee that the generated branches reproduce the reference topology or spatial arrangement.

\subsection{Limitations}
A primary limitation of the proposed method is its dependence on the quality of the initial TreeQSM reconstruction. TreeQSM is sensitive to point cloud quality and parameter settings, and may produce topological errors, particularly for occluded or thin branches. Such errors can propagate into anchor selection and graph encoding, degrading completion quality. To mitigate this issue, we perform a parameter search for each tree based on point cloud density and branch size distribution, following the recommended parameter ranges for TreeQSM and run the reconstruction multiple times to assess stability. These measures help ensure that the partial QSMs provide a reliable structural basis for the completion process.

A second limitation is that our evaluation reference relies on multi-station TreeQSM reconstructions rather than destructively measured woody volume. While multi-station scanning significantly reduces occlusion, it cannot eliminate inherent QSM reconstruction uncertainties, parameter sensitivity, or the known overestimation of small branches. Consequently, the sub-percent volume errors reported here quantify the model's structural consistency relative to the digital references, and should not be interpreted as absolute accuracy against physical tree volume.

The paired experiments use reference-aligned partial QSMs. This construction establishes consistent cylinder identifiers but also replaces the retained regions with reference cylinder geometry. The reported results therefore evaluate missing-structure completion under an aligned representation and do not directly measure performance on independently fitted, unaligned partial QSMs.

\section{Conclusion} \label{sec:7}
% ============================================================

This paper presents a volume and skeleton-aware QSM completion framework for recovering complete tree woody structures from incomplete TLS-derived observations. By moving completion from the raw point cloud domain to the QSM cylinder graph domain, the proposed method directly models the structural elements needed for tree architecture and woody volume estimation. 

The key contributions of this work are threefold. First, we formulate incomplete tree reconstruction as a graph completion problem, leveraging the explicit skeleton and geometry of QSM cylinders. A multi-relation graph encoder with parent, child, and intra-branch adjacency captures the hierarchical branching structure, enabling skeleton-aware feature learning. Second, we develop an anchor-conditioned dual-mode decoder that separately models terminal branch continuation and missing branch-start generation, recovering both branch extensions and missing side branches. Third, we introduce volume-aware learning with total and level-wise volume losses, complemented by constrained inference that uses predicted volume budgets to control the acceptance of generated cylinders.

Extensive experiments on both simulated DART TLS data and real Hong Kong urban TLS trees demonstrate the effectiveness of the proposed method. In terms of volumetric accuracy, the method achieves markedly lower total volume relative error than point cloud completion baselines, while its level-wise volume error remains small, confirming that volume is correctly distributed across branch orders. The volume relative bias is close to zero, indicating that the method is unbiased and neither systematically over-generates nor under-generates volume. Regarding geometric fidelity, the skeleton Chamfer distance is substantially lower than that of point completion baselines, demonstrating that the completed skeleton maintains high geometric similarity to the reference. These results indicate that the proposed method can closely reproduce the volume and skeleton geometry of the selected reference QSMs under the evaluated species and acquisition conditions.

Ablation experiments support the role of the evaluated volume-supervision terms in reference-volume agreement and of axis smoothness in centreline geometry. The acquisition sensitivity experiment identifies limitations under coarse angular sampling in the tested DART scenes, but does not establish a universal resolution requirement for real TLS surveys.

%The proposed method addresses a fundamental limitation of TLS-based forest inventory: the incompleteness of single-station acquisitions. By recovering missing branches and volume in a structurally explicit manner, the method enhances the utility of TLS for forest ecology, urban tree management, and carbon stock estimation. 
%The code and trained models will be released to facilitate reproducibility and practical adoption.

Future work will focus on two directions, which include (1) developing an end-to-end framework that learns directly from point clouds to QSM completion, bypassing the intermediate TreeQSM reconstruction step; and (2) adapting the framework to handle point clouds acquired by mobile mapping systems (MMS), which typically exhibit more severe occlusion and lower point density compared to TLS. This would enable more efficient and flexible acquisition devices, significantly reducing field survey time and expanding the applicability of QSM-based tree structure completion to large-scale urban and forest inventories.

% ============================================================
%\section*{Data availability}
%\textcolor{red}{TLS data, DART simulations, QSM files, and trained models will be publicly available}
%
%\section*{Code availability}
%\textcolor{red}{Code with repository link}

\section*{CRediT authorship contribution statement}
\textbf{Ge Jiang}: Methodology, Software, Data curation, Validation, Formal analysis, Visualization, Writing -- original draft. \textbf{Tiangang Yin}: Conceptualization, Methodology, Resources, Supervision, Project administration, Writing -- review \& editing. \textbf{Wei Yao}: Conceptualization, Supervision, Funding acquisition, Writing -- review \& editing. \textbf{Wai Yeung Yan}: Conceptualization, Supervision, Formal analysis, Writing -- review \& editing. \textbf{Qi Chen}: Investigation, Data curation, Project administration. \textbf{Jinhong Wang}: Investigation, Data curation, Project administration. \textbf{Qihao Weng}: Conceptualization, Methodology, Resources, Funding acquisition, Project administration, Writing -- review \& editing.

\section*{Declaration of competing interest}
The authors declare that they have no known competing financial interests or personal relationships that could have appeared to influence the work reported in this paper.

\section*{Acknowledgements}
This work was supported by the Hong Kong RGC General Research Fund (Grant No. 15215023), the Global STEM Professorship of Hong Kong SAR Government (P0039329), and the Hong Kong Polytechnic University (Project ID: WZ82). The authors thank Kai Him Otto Yu, Shiliang Fu, and Kaichao Wang for their assistance in TLS field data collection, measurements, and data preprocessing.

% Numbered list
% Use the style of numbering in square brackets.
% If nothing is used, default style will be taken.
%\begin{enumerate}[a)]
%\item 
%\item 
%\item 
%\end{enumerate}  

% Unnumbered list
%\begin{itemize}
%\item 
%\item 
%\item 
%\end{itemize}  

% Description list
%\begin{description}
%\item[]
%\item[] 
%\item[] 
%\end{description}  

%\clearpage %%Remove this from your manuscript

% Uncomment and use as the case may be
%\begin{theorem} 
%\end{theorem}

% Uncomment and use as the case may be
%\begin{lemma} 
%\end{lemma}

%% The Appendices part is started with the command \appendix;
%% appendix sections are then done as normal sections
%% \appendix

% To print the credit authorship contribution details
\printcredits

%% Loading bibliography style file
%\bibliographystyle{model1-num-names}
\bibliographystyle{cas-model2-names}

% Loading bibliography database
\bibliography{cas-refs}

% Biography
%\bio{}
% Here goes the biography details.
%\endbio

%\bio{pic1}
% Here goes the biography details.
%\endbio

\end{document}